\pdfoutput=1

\documentclass[11pt]{article}

\usepackage[final]{acl}

\usepackage{times}
\usepackage{latexsym}
\usepackage{enumitem}
\usepackage{amsmath}
\usepackage{multirow}
\usepackage{booktabs}
\usepackage{xcolor}
\usepackage{hhline}

\usepackage[T1]{fontenc}
\usepackage{subcaption}

\usepackage[utf8]{inputenc}

\usepackage{microtype}

\usepackage{inconsolata}

\usepackage{graphicx}
\usepackage{amssymb}
\usepackage{bbm}
\usepackage{dsfont}

\usepackage{hyperref}
\hypersetup{
    colorlinks=true,
    linkcolor=blue,
    citecolor=blue,
    filecolor=magenta,
    urlcolor=blue,
}

\usepackage{colortbl}
\definecolor{em}{gray}{0.9}
\newcommand{\cem}{\cellcolor{em}}

\title{Improving LLM Reasoning through Interpretable Role-Playing Steering}

\author{
  \textbf{Anyi Wang\textsuperscript{1}},
  \textbf{Dong Shu\textsuperscript{2}},
  \textbf{Yifan Wang\textsuperscript{1}},
  \textbf{Yunpu Ma\textsuperscript{1,3}\textsuperscript{†}},
  \textbf{Mengnan Du\textsuperscript{4}\textsuperscript{†}}
\\
\textsuperscript{1}Center for Information and Language Processing, LMU Munich,
\textsuperscript{2}Northwestern University\\
\textsuperscript{3}Munich Center for Machine Learning (MCML)
\textsuperscript{4}New Jersey Institute of Technology\\
\texttt{\{anyi.wang,\,yifan.wang\}@campus.lmu.de, dongshu2024@u.northwestern.edu},\\ 
\texttt{cognitive.yunpu@gmail.com, mengnan.du@njit.edu}
\\
\textsuperscript{†}Corresponding authors
}

\begin{document}
\maketitle
\begin{abstract}

Role-playing has emerged as an effective technique for enhancing the reasoning capabilities of large language models (LLMs). However, existing methods primarily rely on prompt engineering, which often lacks stability and interpretability. In this paper, we introduce \textbf{Sparse Autoencoder Role-Playing Steering (SRPS)}, a novel framework that identifies and manipulates internal model features associated with role-playing behavior. Our approach extracts latent representations from role-play prompts, selects the most relevant features based on activation patterns, and constructs a steering vector that can be injected into the model’s residual stream with controllable intensity. Our method enables fine-grained control over role-specific behavior and offers insights into how role information influences internal model activations. Extensive experiments across various reasoning benchmarks and model sizes demonstrate consistent performance gains. Notably, in the zero-shot chain-of-thought (CoT) setting, the accuracy of Llama3.1-8B on CSQA improves from 31.86\% to 39.80\%, while Gemma2-9B on SVAMP increases from 37.50\% to 45.10\%. These results highlight the potential of SRPS to enhance reasoning ability in LLMs, providing better interpretability and stability compared to traditional prompt-based role-playing. 

\end{abstract}

\section{Introduction}

Large language models (LLMs) have demonstrated remarkable capabilities across a wide range of natural language understanding and generation tasks, including question answering \cite{allemang2024increasing}, text classification \cite{zhang2024pushing}, summarization \cite{zhang2024systematic}, and dialogue systems \cite{yi2024survey}. These models serve as foundational components for numerous downstream applications, benefiting from their strong generalization abilities and sophisticated understanding of context.

One increasingly important application of LLMs is role-playing, in which a model is instructed to behave in a manner consistent with a specified character, persona, or domain expert \cite{shanahan2023role}. In role-playing, the model is expected to adapt its linguistic style, domain knowledge, and thought process to reflect the perspective of a particular identity. A specific example of such a role-play prompt could be: ``\emph{As a highly qualified mathematics teacher, you excel at solving problems systematically and explaining solutions with clarity. Please solve the following problem:}''. This approach has been shown to effectively enhance the model’s performance in reasoning tasks, since the assumed role often helps shape the model's reasoning patterns in useful ways \cite{kong2023better}.

\begin{figure*}
    \centering
    \includegraphics[width=0.98\textwidth]{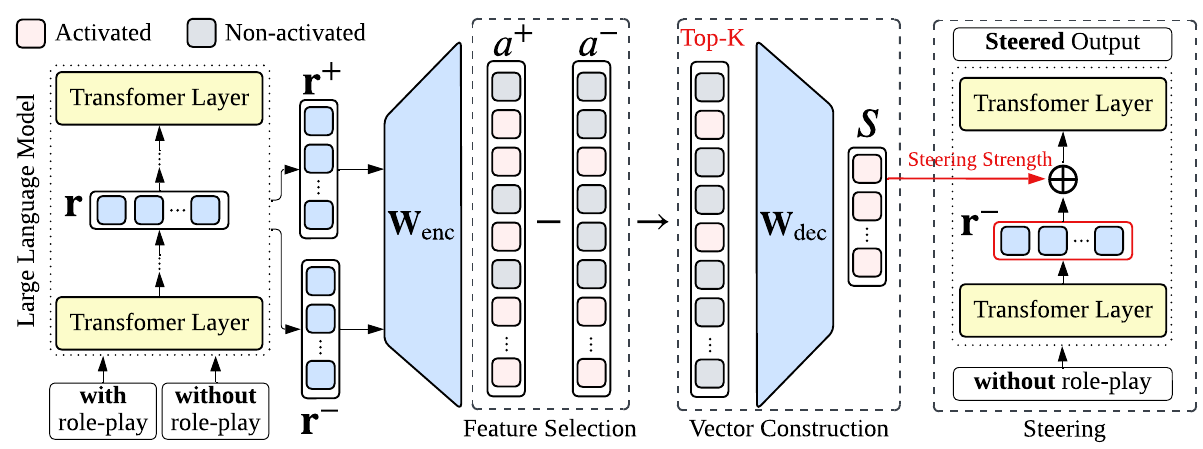}
    \caption{Overview of the SRPS Framework. The LLM takes two types of inputs: one with a role-play prompt and one without. The representation with the role-play prompt is denoted as $\mathbf{r}^+$, while $\mathbf{r}^-$ denotes the representation without the role-play prompt. Both representations are passed through an encoder and perform feature selection that identifies the Top-K steering vectors. These selected vectors are then passed through a decoder to construct the final steering shift vector $\mathbf{s}$. This vector is injected into $\mathbf{r}^-$ to steer the model’s output toward the desired role behavior.}
    \label{fig:figure1}
\end{figure*}

Despite the promise of role-playing techniques, current approaches to leveraging them remain limited in several key aspects. First, most existing methods rely heavily on prompt engineering, often by prepending a description of the desired role to the input \cite{kong2023better,10.1145/3688864.3689149}. Such methods are highly unstable, even minor changes in punctuation, word choice, or formatting can lead to significant variations in performance \cite{shu-etal-2024-dont, mizrahi-etal-2024-state}. This instability arises because LLMs are sensitive to prompt phrasing, and prompts that seem reasonable to humans are not necessarily effective for LLMs \cite{liu2024gpt, liu2021p}. Second, these approaches operate entirely at the input level, offering little interpretability or control over the model’s internal reasoning process. As a result, it is difficult to understand how role information influences model behavior, or to systematically steer the model toward desired behaviors. Third, prior studies on role-playing have primarily evaluated its effectiveness under zero-shot and zero-shot CoT settings \cite{Wei2022ChainOT, Kojima2022LargeLM}, leaving its performance in more practical one-shot and few-shot scenarios largely unexplored \cite{kong2023better,10.1145/3688864.3689149}. 


To address the limitations of prompt-based role-playing, we propose a stable and interpretable steering approach, \textbf{Sparse Autoencoder Role-Playing Steering (SRPS)}. We use a pretrained Sparse Autoencoder (SAE) to analyze the internal activations of an LLM during role-play prompting. We identify the top‑$k$ features that are most strongly and frequently activated by role-play prompts, and use them to construct a steering vector that captures the behavioral shift induced by role-playing. At inference time, this vector is added to the model’s residual stream with a tunable scaling factor, enabling stable steering of the model’s internal state toward the desired role-based reasoning behavior. To better understand why the top‑$k$ features influence the model’s reasoning process, we use Neuronpedia \cite{neuronpedia} to interpret their semantic functions, offering insight into how role-playing enhances model performance. Finally, we examine accuracy variation across different sets of \(k\) features, demonstrating the effectiveness of our feature ranking method.
Extensive experiments show that our steering mechanism consistently enhances the model's reasoning ability and achieves performance comparable to prompt-based role-playing in most cases. Our contributions can be summarized as follows:
\begin{itemize}[leftmargin=*]\setlength\itemsep{-0.3em}
    \item We propose SRPS, a fine-grained and stable steering method that modulates the internal behavior of LLMs. 

    \item We reveal the underlying mechanism of role-playing effectiveness using features extracted from pretrained SAEs. 
    

    \item Experiments across multiple models and reasoning benchmarks indicate that the proposed SRPS consistently improves model performance and performs as well as prompt-based approaches.

\end{itemize}

\section{Related Work}
\subsection{Sparse Autoencoders}
Steering methods have emerged as effective approaches for controlling and interpreting the behavior of LLMs \cite{wu2024reft, templeton2024scaling, arditi2024refusal, 10.5555/3692070.3693926, bi2024visual,li2025featurecot,he2025saessv}. SAEs \cite{shu2025surveysparseautoencodersinterpreting} have further strengthened this direction by offering a powerful tool for uncovering interpretable structures within LLMs. \citet{cunningham2023sparse} demonstrate that SAEs can extract highly interpretable and monosemantic features from LLM activations, enabling fine-grained analysis of model behavior. Building upon this, \citet{bricken2023monosemanticity} apply dictionary learning techniques to decompose LLM representations into semantically meaningful components, facilitating a deeper understanding of their internal mechanisms. To enhance the reconstruction fidelity of SAEs without compromising interpretability, \citet{rajamanoharan2024jumping} introduce JumpReLU SAEs, which utilize a discontinuous activation function to achieve state-of-the-art performance on large models like Gemma2-9B \cite{team2024gemma}. Furthermore, \citet{gao2024scaling} explore the scalability of SAEs, training models with up to 16 million latents on GPT-4 \cite{achiam2023gpt} activations and establishing clear scaling laws that balance sparsity and reconstruction accuracy. These advancements highlight the effectiveness of SAEs in interpreting and analyzing the complex behaviors of LLMs.

\subsection{Role-Play Prompting}

Role-play prompting has been explored as a technique to enhance the reasoning capabilities of LLMs by assigning them specific personas or roles \cite{shanahan2023role}. \citet{kong2023better} introduce a structured two-stage role-play prompting procedure. In stage 1, a role-setting prompt is constructed and multiple role-feedback responses are sampled; in stage 2, both the role-setting prompt and the selected optimal role-feedback response are provided to the model for answer generation, making the role-play context more immersive. Extensive experiments demonstrate that the strategy consistently outperforms standard zero-shot approach across various reasoning benchmarks. However, the effectiveness of role-play prompting is not universal. \citet{10.1145/3688864.3689149} critically examine its application in mathematical reasoning tasks, finding that directly adding role-play prompts before questions does not always enhance model performance and may sometimes even degrade it. These studies collectively highlight the potential and limitations of role-play prompting, emphasizing the need for general strategies that can reliably and consistently improve LLM reasoning performance across diverse tasks.

\subsection{Chain-of-Thought Reasoning}

The idea of CoT reasoning is to make the model mimic the human thought process when solving complicated reasoning tasks \cite{Wei2022ChainOT}. It enables the model to break down the problem into several intermediate steps and solve each one before arriving at the final answer. This approach has been shown to notably enhance the model’s reasoning ability and significantly improve its performance on arithmetic, symbolic, and commonsense reasoning tasks. In few-shot CoT prompting \cite{Wei2022ChainOT}, step-by-step reasoning process is demonstrated in the exemplars and fed into the model as input. And in zero-shot CoT \cite{Kojima2022LargeLM}, CoT reasoning can be elicited from LLMs by simply adding ``Let’s think step by step'' to the prompt. This requires no fine-tuning or task specific conditioning, and substantially outperforms standard zero-shot learning and sometimes even few-shot learning on a wide range of tasks.

\section{Proposed Steering Framework}

In this section, we present our proposed SAE-based role-playing steering (SRPS) framework. As shown in Figure \ref{fig:figure1}, we begin by constructing role-playing prompt sets across different domains, which serve as inputs for activation extraction. Next, we identify the top-$k$ features that are most relevant to role-playing by analyzing their activation strength and frequency on role-play prompts. Based on the selected features, we compute a steering shift vector and inject it into the model’s residual stream, scaled by a tunable parameter that controls the strength of steering. 

\subsection{Extracting Role-Play Activations}

We construct \( N \) sample pairs from the training set for every dataset, each consisting of a positive and a negative sample. The positive sample includes a question prepended with a role-play prompt, while the negative sample is the same question without the role-play instruction. To avoid overfitting to the surface form of a single instruction, where the model memorizes its literal meaning instead of learning a generalizable behavior vector \cite{He2025SAIFAS}, we design five semantically diverse prompt variants for each role.

For each prompt pair, we extract the residual stream representations of the samples and compute their corresponding latent activations using the pretrained SAE. We then calculate the mean activation of each sample by averaging over all tokens, which we use to identify features associated with role-playing. When averaging activations, we exclude punctuation, stop words, and the Beginning Of Sequence (BOS) token. As these tokens frequently occur in role-play prompts but carry limited semantic value, and including them could cause the steering vector to rely on spurious patterns, potentially degrading steering effectiveness.

\subsection{Role-Relevant Feature Selection}




To identify features that are most relevant to role-play behavior, we compute the mean SAE latent activation differences across each sample pair. Specifically, \( a_{ij}^{+} \) represents the activation of the \( i \)-th SAE feature for the \( j \)-th sample with a role-play prefix, while \( a_{ij}^{-} \) denotes the corresponding activation for the same sample without role-play prompt. The average activation difference for each feature across \( N \) sample pairs is computed as:
\begin{equation}
    \mu_i = \frac{1}{N} \sum_{j=1}^{N} (a_{ij}^{+} - a_{ij}^{-}).
    \label{eq:mean-activation-diff}
\end{equation}


In addition to activation strength, we evaluate the frequency with which each feature is activated above a threshold \( \theta \) across \( N \) sample pairs. With \( \mathbbm{1}(\cdot) \) being the indicator function, we define the activation frequency difference between samples with and without role-play prompts as:
\begin{align}
    \delta_i &= f_i^{+} - f_i^{-} \nonumber \\
    &= \frac{1}{N} \sum_{j=1}^{N} \mathbbm{1}(a_{ij}^{+} > \theta) - \frac{1}{N} \sum_{j=1}^{N} \mathbbm{1}(a_{ij}^{-} > \theta),
    \label{eq:freq-diff-combined}
\end{align}
where \( f_i^{+} \) and \( f_i^{-} \) represent the activation frequency of the \( i \)-th feature in samples with and without role-play prompts respectively.
Finally, to integrate both activation strength and frequency, we define a sensitivity score for each feature as:
\begin{equation}
    I_i = \mu_i + \beta \cdot \delta_i,
    \label{eq:sensitivity-score}
\end{equation}
where \( \beta \) is a tunable hyperparameter that balances the influence of the two components.

The sensitivity score quantifies the extent to which a feature's activation shifts between samples with and without role-play prompts, capturing both strength and frequency differences. A higher score indicates that the feature is more strongly influenced by role-playing and is therefore more relevant for steering. The top-\( k \) features with the highest sensitivity scores are then selected for constructing the steering shift vector.

\subsection{Steering Vector Construction}

After selecting the top-$k$ role-relevant features, we retrieve their steering vectors from the SAE decoder \( W_{\mathrm{dec}} \in \mathbb{R}^{d \times h} \), where \( d \) is the number of features and \( h \) is the dimensionality of the residual stream. 
The steering vector corresponding to the \( i \)-th selected feature is represented as \( \mathbf{v}_i = W_{\mathrm{dec}}[i,:] \).

As sentence-level instructions contain rich and complex meanings spread across multiple features, using a single vector is often insufficient for effective steering~\cite{He2025SAIFAS}. Therefore, we use a set of vectors to better capture the full effect of role-playing. 
We define the overall steering shift as a weighted combination of the top-$k$ steering vectors, where the weight for each selected feature corresponds to its mean activation \( \alpha_i \) across \( N \) positive samples. The final steering shift vector \( \mathbf{s} \in \mathbb{R}^h \) is computed as:
\begin{equation}
\mathbf{s} = \sum_{i=1}^{k} \alpha_i \cdot \mathbf{v}_i.
\label{eq:steering-shift}
\end{equation}
To apply the computed shift vector, the steering signal is injected into the residual stream representation of the input over the last token at layer \( l \), thereby adjusting the model’s internal activations before decoding. The original residual stream representation at this position is represented as \( \mathbf{r} \), and \( \lambda \) is a tunable scaling factor that controls the steering strength. The updated residual stream representation is computed as:
\begin{equation}
\mathbf{r}_{\mathrm{new}} = \mathbf{r} + \lambda \cdot \mathbf{s}.
\label{eq:residual-update}
\end{equation}
Finally, to maintain the original capabilities of the LLM and ensure stability during the steering process, we follow the normalization strategy proposed by \citet{Liu2023IncontextVM} and apply it to the updated representations:
\begin{equation}
\mathbf{r}_{\mathrm{new}} := \mathbf{r}_{\mathrm{new}} \cdot \frac{\|\mathbf{r}\|_2}{\|\mathbf{r}_{\mathrm{new}}\|_2}.
\label{eq:residual-norm}
\end{equation}

\begin{table*}[ht]
\resizebox{\textwidth}{!}{%
\begin{tabular}{|ll|lll|lll|lll|}
\toprule
\toprule
\multicolumn{2}{c|}{Benchmark}                                                    & \multicolumn{3}{c|}{GSM8K}                                                 & \multicolumn{3}{c|}{SVAMP}                                                 & \multicolumn{3}{c}{CSQA}                                                          \\
\midrule
\multicolumn{2}{c|}{Evaluation Setting}                                           & \multicolumn{1}{l|}{4-shot} & \multicolumn{1}{l|}{1-shot} & 0-shot         & \multicolumn{1}{l|}{4-shot} & \multicolumn{1}{l|}{1-shot} & 0-shot         & \multicolumn{1}{l|}{4-shot} & \multicolumn{1}{l|}{1-shot}         & \multicolumn{1}{l}{0-shot}        \\
\midrule
\multicolumn{1}{c|}{\multirow{4}{*}{Llama3.1-8B}} & original prompting            & \multicolumn{1}{l|}{52.84}  & \multicolumn{1}{l|}{42.68}  & 40.26          & \multicolumn{1}{l|}{67.60}  & \multicolumn{1}{l|}{60.30}  & 59.80          & \multicolumn{1}{l|}{72.24}  & \multicolumn{1}{l|}{66.75}          & \multicolumn{1}{l}{31.86}         \\

\multicolumn{1}{c|}{}                             & role-play prompting   & \multicolumn{1}{l|}{54.51 \textcolor{green!60!black}{↑}}  & \multicolumn{1}{l|}{41.85 \textcolor{red}{↓}}  & 38.13 \textcolor{red}{↓}          & \multicolumn{1}{l|}{65.40 \textcolor{red}{↓}}  & \multicolumn{1}{l|}{55.90 \textcolor{red}{↓}}  & 54.60 \textcolor{red}{↓}          & \multicolumn{1}{l|}{\textbf{73.71 \textcolor{green!60!black}{↑}}}  & \multicolumn{1}{l|}{65.19 \textcolor{red}{↓}}          & \multicolumn{1}{l}{27.27 \textcolor{red}{↓}}          \\
\multicolumn{1}{c|}{}                             & \cem{\textbf{SAE-based steering}}                & \multicolumn{1}{l|}{\cem{\textbf{54.66 \textcolor{green!60!black}{↑}}}}  & \multicolumn{1}{l|}{\cem\textbf{43.75 \textcolor{green!60!black}{↑}}}  & \multicolumn{1}{l|}{\cem{\textbf{44.66 \textcolor{green!60!black}{↑}}}}          & \multicolumn{1}{l|}{\cem{\textbf{68.20 \textcolor{green!60!black}{↑}}}} & \multicolumn{1}{l|}{\cem{\textbf{60.60 \textcolor{green!60!black}{↑}}}}  & \multicolumn{1}{l|}{\cem{\textbf{63.70 \textcolor{green!60!black}{↑}}}}         & \multicolumn{1}{l|}{\cem{72.65 \textcolor{green!60!black}{↑}}}  & \multicolumn{1}{l|}{\cem{\textbf{69.78 \textcolor{green!60!black}{↑}}}}        & \multicolumn{1}{l}{\cem{\textbf{39.80 \textcolor{green!60!black}{↑}}}}     \\
\midrule
\multicolumn{1}{c|}{\multirow{4}{*}{Gemma2-2B}}   & original prompting            & \multicolumn{1}{l|}{25.85}  & \multicolumn{1}{l|}{14.63}  & 18.35          & \multicolumn{1}{l|}{46.70}  & \multicolumn{1}{l|}{36.60}  & 31.60          & \multicolumn{1}{l|}{\textbf{63.06}}  & \multicolumn{1}{l|}{56.43}          & \multicolumn{1}{l}{17.20}          \\

\multicolumn{1}{c|}{}                             & role-play prompting    & \multicolumn{1}{l|}{\textbf{27.90 \textcolor{green!60!black}{↑}}}  & \multicolumn{1}{l|}{\textbf{18.20 \textcolor{green!60!black}{↑}}}  & \multicolumn{1}{l|}{\textbf{24.26 \textcolor{green!60!black}{↑}}}          & \multicolumn{1}{l|}{45.50 \textcolor{red}{↓}}  & \multicolumn{1}{l|}{\textbf{39.20} \textcolor{green!60!black}{↑}}  & \multicolumn{1}{l|}{\textbf{55.10 \textcolor{green!60!black}{↑}}}          & \multicolumn{1}{l|}{60.11 \textcolor{red}{↓}}  & \multicolumn{1}{l|}{54.22 \textcolor{red}{↓}}          & \multicolumn{1}{l}{\textbf{25.88 \textcolor{green!60!black}{↑}}} \\
\multicolumn{1}{c|}{}                             &\cem{\textbf{SAE-based steering}}                 & \multicolumn{1}{l|}{\cem{26.61 \textcolor{green!60!black}{↑}}}  & \multicolumn{1}{l|}{\cem{17.44 \textcolor{green!60!black}{↑}}}  & \multicolumn{1}{l|}{\cem{22.06 \textcolor{green!60!black}{↑}}}          & \multicolumn{1}{l|}{\cem{\textbf{47.40 \textcolor{green!60!black}{↑}}}}  & \multicolumn{1}{l|}{\cem{38.10 \textcolor{green!60!black}{↑}}}  & \multicolumn{1}{l|}{\cem{34.20 \textcolor{green!60!black}{↑}}}          & \multicolumn{1}{l|}{\cem{\textbf{63.06 \textcolor{gray}{\textrightarrow}}}}  & \multicolumn{1}{l|}{\cem{\textbf{59.54 \textcolor{green!60!black}{↑}}}}          & \multicolumn{1}{l}{\cem{20.56 \textcolor{green!60!black}{↑}}}          \\
\midrule
\multicolumn{1}{c|}{\multirow{4}{*}{Gemma2-9B}}   & original prompting            & \multicolumn{1}{l|}{66.57}  & \multicolumn{1}{l|}{61.64}  & 39.12          & \multicolumn{1}{l|}{81.20}  & \multicolumn{1}{l|}{79.20}  & 37.50          & \multicolumn{1}{l|}{79.36}  & \multicolumn{1}{l|}{71.99}          & \multicolumn{1}{l}{41.69}          \\

\multicolumn{1}{c|}{}                             & role-play prompting    & \multicolumn{1}{l|}{65.50 \textcolor{red}{↓}}  & \multicolumn{1}{l|}{59.51 \textcolor{red}{↓}}  & \textbf{49.73 \textcolor{green!60!black}{↑}} & \multicolumn{1}{l|}{79.40 \textcolor{red}{↓}}  & \multicolumn{1}{l|}{75.70 \textcolor{red}{↓}}  & \textbf{75.30 \textcolor{green!60!black}{↑}} & \multicolumn{1}{l|}{78.13 \textcolor{red}{↓}}  & \multicolumn{1}{l|}{\textbf{73.46 \textcolor{green!60!black}{↑}}} & \multicolumn{1}{l}{46.44 \textcolor{green!60!black}{↑}}          \\
\multicolumn{1}{c|}{}                             & \cem{\textbf{SAE-based steering}}                 & \multicolumn{1}{l|}{\cem{\textbf{66.79 \textcolor{green!60!black}{↑}}}}  & \multicolumn{1}{l|}{\cem{\textbf{61.87 \textcolor{green!60!black}{↑}}}}  & \multicolumn{1}{l|}{\cem{39.88 \textcolor{green!60!black}{↑}}}          & \multicolumn{1}{l|}{\cem{\textbf{81.30 \textcolor{green!60!black}{↑}}}}  & \multicolumn{1}{l|}{\cem{\textbf{79.80 \textcolor{green!60!black}{↑}}}}  & \multicolumn{1}{l|}{\cem{45.10 \textcolor{green!60!black}{↑}}}          & \multicolumn{1}{l|}{\cem{\textbf{80.02 \textcolor{green!60!black}{↑}}}}  & \multicolumn{1}{l|}{\cem{71.99 \textcolor{gray}{\textrightarrow}}}          & \multicolumn{1}{l}{\cem{\textbf{48.08 \textcolor{green!60!black}{↑}}}} \\
\bottomrule[1pt]
\bottomrule[1pt]
\end{tabular}%
}
\caption{Accuracy comparison of models with different methods on each dataset. Green, red and gray arrows indicate performance changes against the original prompting. The best result in each group is highlighted in \textbf{bold}.}
\label{tab:results}
\end{table*}

\section{Experiments}
In this section, we conduct a series of experiments to evaluate the effectiveness and interpretability of our role-playing steering method. We aim to answer the following research questions (RQs):
\begin{itemize}[leftmargin=*]\setlength\itemsep{-0.3em}
    \item \textbf{RQ1:} How effectively does our steering method improve model performance? (Section 4.2)

    \item \textbf{RQ2:} How do the selected SAE features explain the model's reasoning enhancement under role-playing? (Section 4.3)

    \item \textbf{RQ3:} How do the features with different rankings affect steering performance? (Section 4.4)
\end{itemize}

\subsection{Experimental Settings}
\noindent\textbf{Datasets and Models.}
We conduct experiments on multiple language models, including Gemma2-2B, Gemma2-9B~\cite{team2024gemma}, and Llama3.1-8B~\cite{grattafiori2024llama}. 
Our evaluation spans three datasets across two reasoning categories: (1) Arithmetic reasoning, including GSM8K \cite{cobbe2021training} and SVAMP \cite{patel2021nlp}; and 
(2) Commonsense reasoning, represented by CSQA \cite{talmor2018commonsenseqa}. Additional dataset details are provided in Appendix~\ref{sec:appendix_datasets}.

\vspace{3pt}\noindent\textbf{Instruction Design.}
We follow the setup of \citet{kong2023better} to design role-play prompts across two reasoning domains. 
For arithmetic reasoning, the LLM is assigned the role of a mathematics teacher, while the user takes the role of a student. For commonsense reasoning, the LLM assumes the role of a contestant in a general knowledge quiz, with the user acting as the moderator. To ensure coherence, we insert a transition sentence between the role-play prompt and the task question. Additionally, to introduce linguistic diversity and avoid prompt overfitting, we design five distinct prompt variants for each role. All role-play prompts are provided in Table \ref{tab:prompt} in Appendix~\ref{sec:appendix_prompts}.

\vspace{3pt}\noindent\textbf{Baselines.} We compare the performance of our proposed steering method against two baselines: the model prompted with the original question, and the model using a prepended role-play prompt. Evaluations are conducted under three settings: few-shot CoT (4-shot), one-shot CoT, and zero-shot CoT. We append ``Let's think step by step'' to the original question to elicit CoT reasoning in LLMs. For one-shot and few-shot exemplars, we follow \citet{Wei2022ChainOT} to construct them (see Table \ref{tab:prompt_math} and \ref{tab:prompt_commonsense} in Appendix \ref{sec:appendix_prompts}). For the prompt-based baseline, we prepend each of the five role-specific prompts designed for steering to the input question respectively, and select the best result among them for comparison. The model accuracy for each role-play prompt, along with the average and standard deviation across the five prompts, is reported in Table~\ref{tab:avarage_prompt} in Appendix~\ref{sec:appendix_prompt_avarage}. 

\vspace{3pt}\noindent\textbf{Implementation Details.}
The number of input sample pairs \( N \) is set to 1,000. We use pretrained SAEs from Gemma Scope \cite{lieberum2024gemma} and Llama Scope \cite{he2024llama} for latent activation extraction. Since deeper layers in LLMs integrate broader contextual information and capture higher-level representations, and SAEs with larger dimensions provide better interpretability of extracted features, we select the later layers and the highest available SAE dimensionality for our experiments. We use the SAE with 131K dimension from layer 25 for Llama3.1-8B, 65K dimension from layer 25 for Gemma2-2B, and 131K dimension from layer 35 for Gemma2-9B. In our experiments, we choose the top 15 SAE latent features with the highest sensitivity scores for steering, as \citet{He2025SAIFAS} demonstrate that the model achieves the optimal performance when steering is performed using 15 latent dimensions. Other hyperparameter choices for all the experimental settings are summarized in Table \ref{tab:hyperparameter} in Appendix~\ref{sec:appendix_hyperparameters}, where we further discuss the influence of different hyperparameters on the steering effect.

\begin{figure*}
    \centering
    \includegraphics[width=0.85\textwidth]{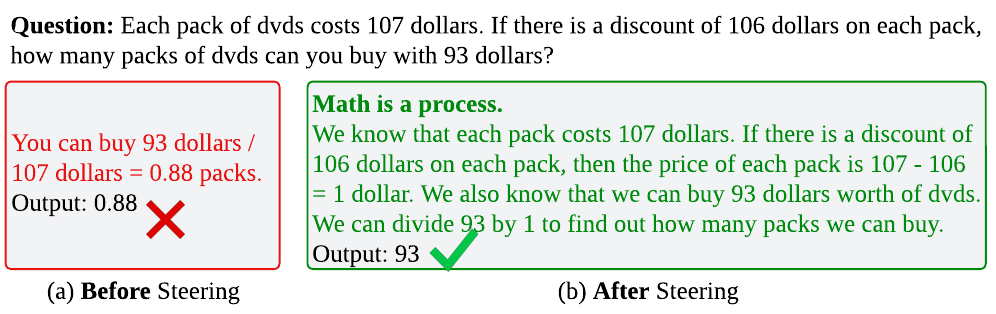}
    \caption{Comparison of model outputs before and after steering, using an example from the SVAMP dataset.}
    \label{fig:svamp_4}
\end{figure*}

\subsection{Benchmark Performance Comparison}

Our main results are presented in Table \ref{tab:results} and are discussed in the following section.

\vspace{3pt}\noindent\textbf{Effectiveness of the Proposed SRPS.} As shown in Table~\ref{tab:results}, our steering approach demonstrates superior performance, outperforming the original prompting in \underline{25 out of 27} evaluation settings and achieving comparable results in the remaining two settings on the CSQA dataset. Compared to the prompt-based role-playing baseline, our method achieves better results in \underline{17 out of 27} evaluation settings. These results highlight the effectiveness of our SAE-based steering mechanism across models and tasks.

\vspace{3pt}\noindent\textbf{Steering Performance Comparison across Different Evaluation Settings.} We conduct experiments under few-shot, one-shot, and zero-shot CoT settings to assess the effectiveness of our steering method. Results reveal that steering is particularly effective in zero-shot CoT, whereas its impact is reduced when few-shot examples are available. For example, on the CSQA dataset with Llama3.1-8B, the accuracy improves slightly under few-shot CoT setting (from 72.24\% to 72.65\%), and by about 3\% (from 66.75\% to 69.78\%) under one-shot CoT setting, while the accuracy increases significantly \underline{from 31.86\% to 39.80\%} in zero-shot CoT. A similar trend is observed with Gemma2-9B, where its zero-shot performance improves \underline{from 37.50\% to 45.10\%} on SVAMP, and \underline{from 41.69\% to 48.08\%} on CSQA. This suggests that steering provides greater benefits when the model lacks demonstrations. In contrast, under few-shot CoT, where model performance is already strong, the potential for further improvement is more limited.

Furthermore, the steering method outperforms prompt-based role-playing in 4 out of 9 zero-shot CoT settings, 6 out of 9 one-shot CoT settings, and 7 out of 9 few-shot CoT settings. This suggests that the steering approach is particularly advantageous under one-shot and few-shot CoT, while both methods perform comparably in zero-shot CoT scenario. However, the performance variation is particularly large using role-play prompts under zero-shot CoT (see Table \ref{tab:avarage_prompt}). Compared to prompting, our steering method offers better interpretability and more stable, controllable behavior, making it a more practical alternative to prompt-based role-playing.

\begin{figure*}[ht]
    \centering
    \begin{subfigure}[t]{0.49\textwidth}
        \centering
        \includegraphics[width=\textwidth]{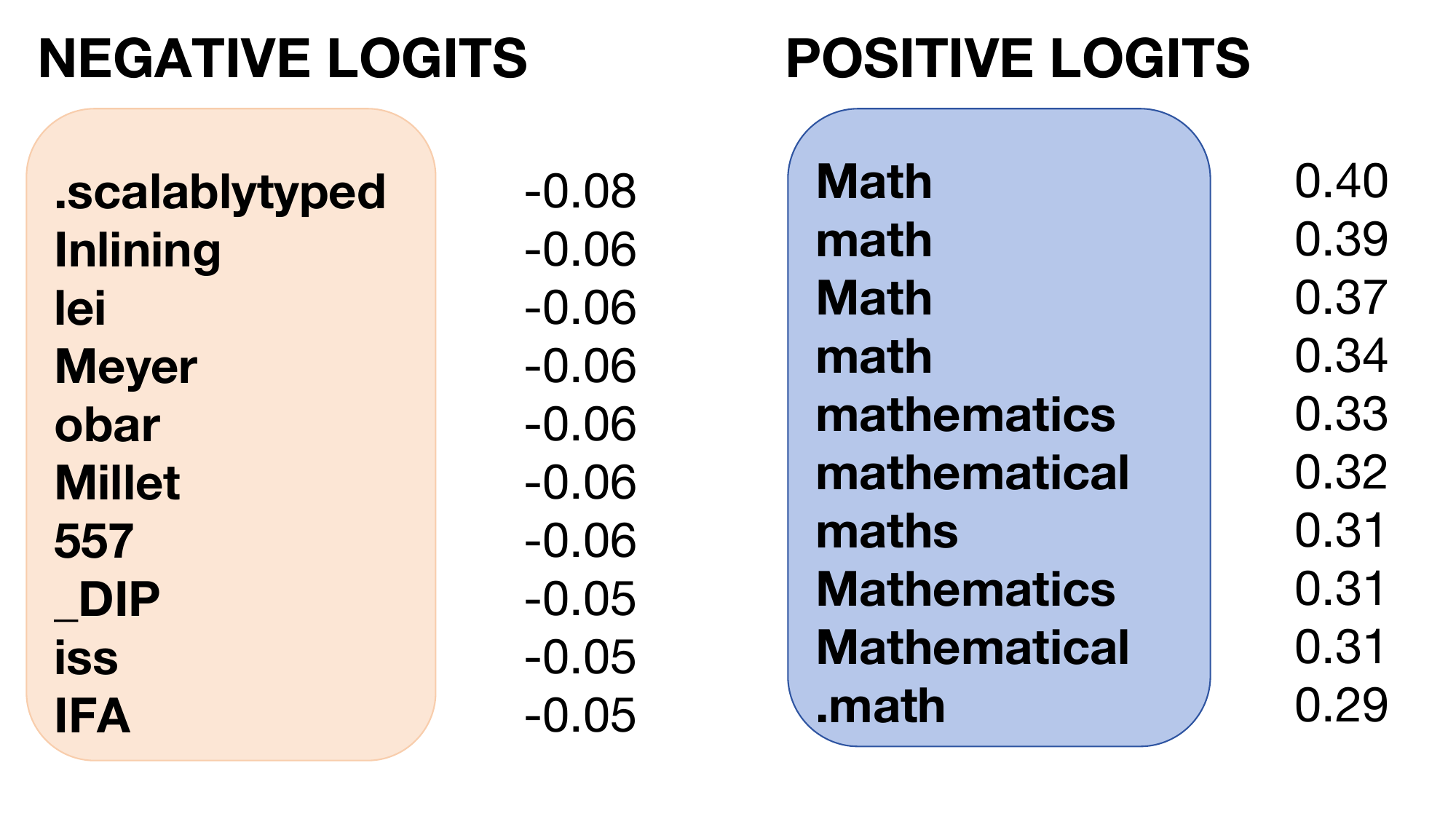}
        \caption{Arithmetic Reasoning}
        \label{fig:math}
    \end{subfigure}%
    \hspace{0.014\textwidth}
    \begin{subfigure}[t]{0.49\textwidth}
        \centering
        \includegraphics[width=\textwidth]{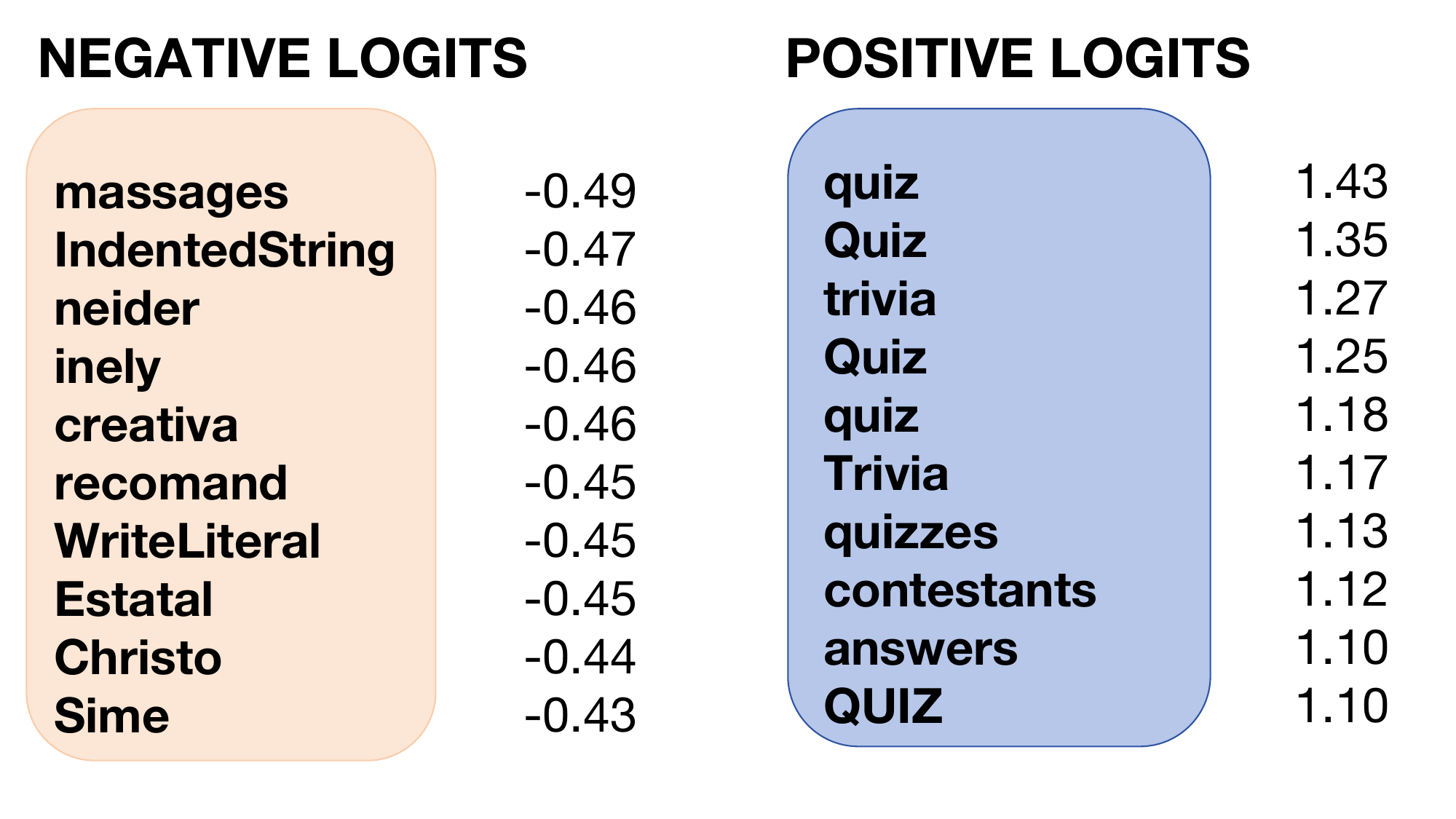}
        \caption{Commonsense Reasoning}
        \label{fig:commonsense}
    \end{subfigure}
    \caption{Top 10 negative and positive output logits of the highest-ranked feature across all evaluation settings in arithmetic and commonsense reasoning tasks. Data sourced from Neuronpedia \cite{neuronpedia}.}
    \label{fig:logits}
\end{figure*}

\begin{table*}[ht]
\resizebox{\textwidth}{!}{%
\begin{tabular}{cc|cc}
\toprule[0.8pt]
\toprule[0.8pt]

\multicolumn{2}{c|}{\textbf{Arithmetic Reasoning}}                                                                   & \multicolumn{2}{c}{\textbf{Commonsense Reasoning}}                     \\ \midrule
\multicolumn{1}{c|}{\textbf{Semantics}}                                          & \textbf{Feature Index}                              & \multicolumn{1}{c|}{\textbf{Semantics}}                          & \textbf{Feature Index} \\ \midrule
\multicolumn{1}{l|}{math, mathematics}                                  & \multicolumn{1}{l|}{74432}                                      & \multicolumn{1}{l|}{quiz}                               & \multicolumn{1}{l}{13980}         \\ 
\multicolumn{1}{l|}{tutoring, teaching}                                 & \multicolumn{1}{l|}{57569, 21959, 47110}                        & \multicolumn{1}{l|}{contestant}                         & \multicolumn{1}{l}{14140}         \\ 
\multicolumn{1}{l|}{problem solving}                                    & \multicolumn{1}{l|}{102613, 126192, 41962} & \multicolumn{1}{l|}{reality}                            & \multicolumn{1}{l}{59617}         \\ 
\multicolumn{1}{l|}{thinking, reasoning, critical, analytical} & \multicolumn{1}{l|}{60715}                                      & \multicolumn{1}{c|}{contest, competition, exams} & \multicolumn{1}{l}{15478, 26757}  \\ 
\multicolumn{1}{l|}{formulas, equations}                                & \multicolumn{1}{l|}{13482}                                      & \multicolumn{1}{l|}{think, focus}                       & \multicolumn{1}{l}{42261}         \\ 
\multicolumn{1}{l|}{coding, programming}                                & \multicolumn{1}{l|}{117154}                                     & \multicolumn{1}{l|}{host}                               & \multicolumn{1}{l}{1733}          \\ 
\multicolumn{1}{l|}{calculus, differential}                             & \multicolumn{1}{l|}{90027}                                      & \multicolumn{1}{l|}{general}                            & \multicolumn{1}{l}{35525}         \\ 
\bottomrule[1pt]
\bottomrule[1pt]
\end{tabular}%
}
\caption{Task-related features extracted from Layer 35 of Gemma2-9B for arithmetic reasoning and from Layer 25 of Gemma2-2B for commonsense reasoning.}
\label{tab:feature}
\end{table*}

\vspace{3pt}\noindent\textbf{Steering Performance Comparison across Different LLMs.} Among all the evaluated models, Llama3.1-8B exhibits the most significant performance improvement after steering, with the accuracy consistently surpassing the original prompting across all evaluation settings evidently. In contrast, Gemma2-9B appears less sensitive to activation perturbation, showing only modest performance gains in most few-shot and one-shot CoT scenarios, which is likely due to its already strong baseline performance.

Compared to the prompting baseline, our steering method yields higher accuracy in 3 out of 9 experimental settings with Gemma2-2B, 6 out of 9 settings with Gemma2-9B, and 8 out of 9 settings with Llama3.1-8B. This trend suggests that steering offers greater advantages over prompting in larger LLMs.

\subsection{SAE Feature Analysis}

To understand how the selected SAE features enhance model reasoning under role-playing, we conduct a semantic analysis of their activations using Neuronpedia \cite{neuronpedia}. Specifically, we analyze features from Layer 35 of Gemma2-9B for arithmetic reasoning and features from Layer 25 of Gemma2-2B for commonsense reasoning, as shown in Table \ref{tab:feature}. Figure~\ref{fig:logits} presents the top-1 ranked feature across all evaluation settings in each domain, showing the corresponding positive and negative output logits. We found the extracted features align well with the expected domain knowledge for each reasoning task, validating the interpretability and task relevance of our steering mechanism. Figure \ref{fig:svamp_4} illustrates the impact of our steering method on model outputs. See Appendix~\ref{sec: steering_example} for more examples.

\begin{figure*}
    \centering
    \subfloat[GSM8K]{
        \includegraphics[width=0.32\linewidth]{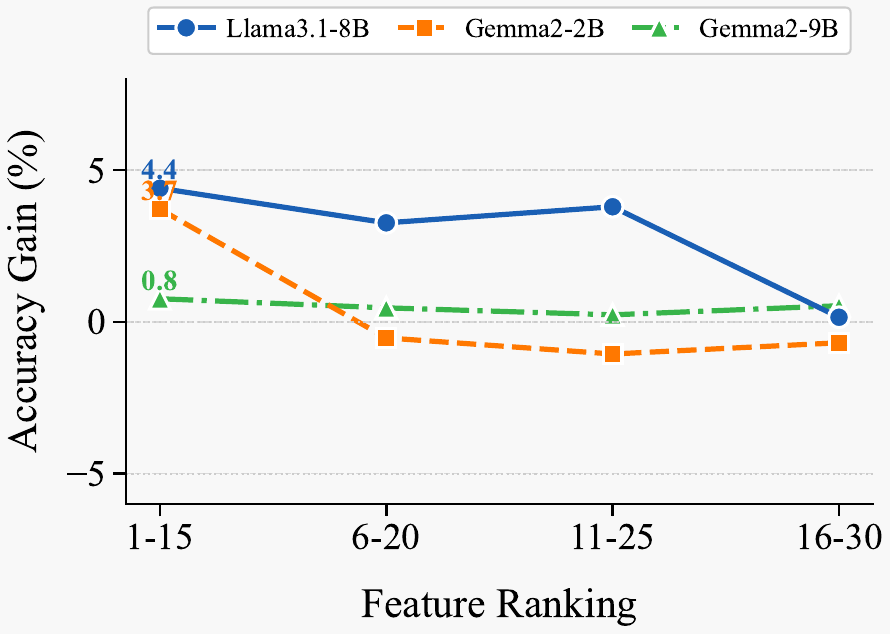}
    }
    \subfloat[SVAMP]{
        \includegraphics[width=0.32\linewidth]{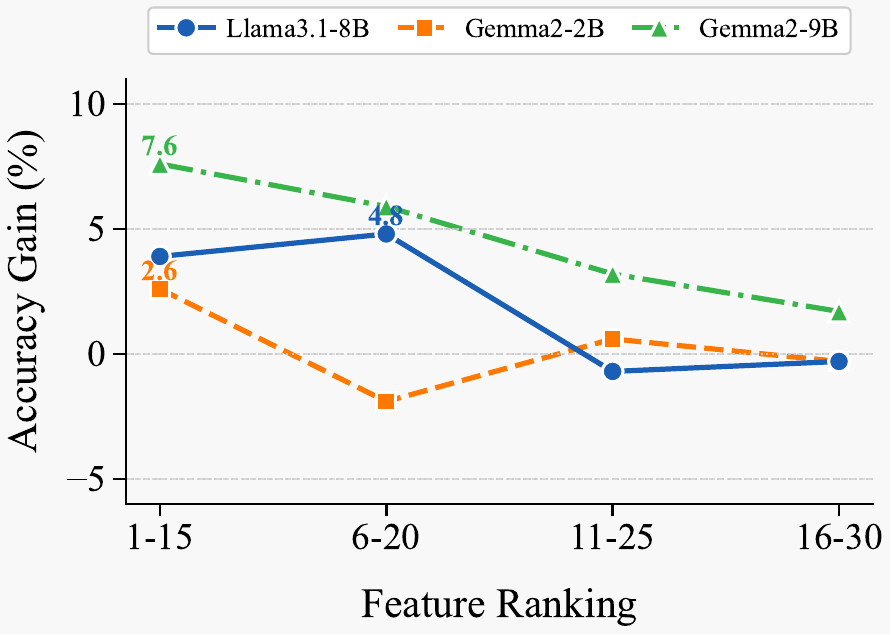}
    }
    \subfloat[CSQA]{
        \includegraphics[width=0.32\linewidth]{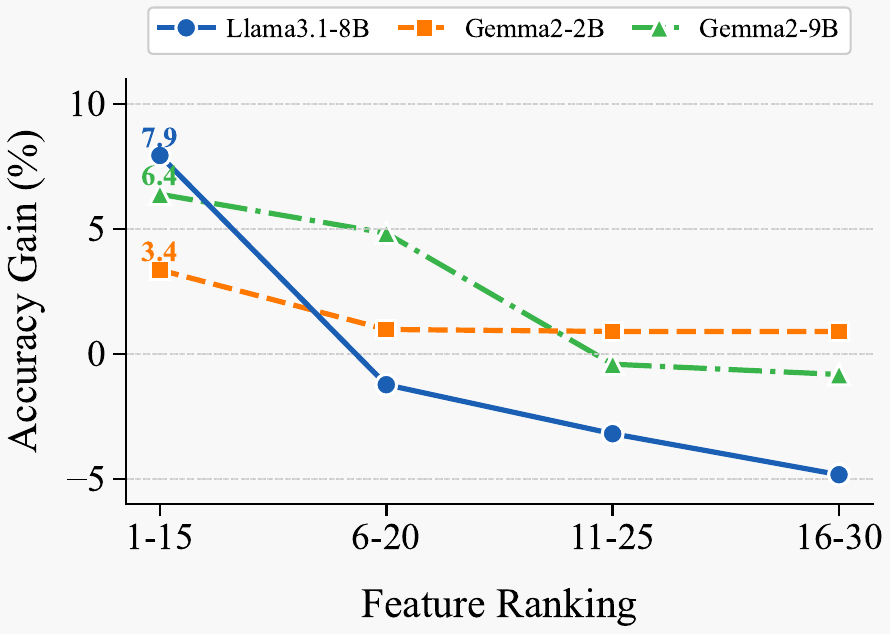}
    }
    \caption{Comparison of accuracy gains over the original prompting under zero-shot CoT setting when steering with different top-\(k\) features.}
    
    \label{fig:accuracy_gain}
\end{figure*}

\vspace{3pt}\noindent\textbf{Arithmetic Reasoning.} The extracted features for arithmetic reasoning exhibit strong semantic alignment with mathematical concepts and problem-solving process. As shown in Figure~\ref{fig:math}, the top-ranked feature is associated with the semantics ``math'', aligning well with the nature of the arithmetic tasks. Additional relevant features are presented in Table~\ref{tab:feature} and analyzed as follows:

\begin{itemize}[leftmargin=*]\setlength\itemsep{-0.3em}
    \item mathematics, formulas, equations, programming, calculus: The semantics of these features are strongly associated with core mathematical terminology. Such alignment is essential for arithmetic reasoning tasks like GSM8K and SVAMP, which require numerical computation and algebraic manipulation.
    
    \item tutoring, teaching: The meaning conveyed by these features suggests that our steering method effectively applies the assigned teacher role to the LLM, thereby guiding the model toward the intended role-consistent behavior. 
    
    \item thinking, reasoning, critical, analytical, logical, problem solving: Analysis of these features reveals that role-play prompts contain elements that are essential for supporting in-depth reasoning in the LLMs. These features likely help trigger step-by-step problem solving, leading to improved overall model performance.
       
\end{itemize}

\vspace{3pt}\noindent\textbf{Commonsense Reasoning.} The features activated through role-playing in commensense reasoning also demonstrate a high degree of semantic relevance to the task. As shown in Figure~\ref{fig:commonsense}, the top-ranked feature is associated with the semantics ``quiz'', which is consistent with the general knowledge contest scenario assigned to the model. Other beneficial features are listed in Table~\ref{tab:feature} and analyzed below:

\begin{itemize}[leftmargin=*]\setlength\itemsep{-0.3em}

    \item quiz, contest, exams, contestant, host: The semantics of these features align closely with the quiz competition context of the role. Framing the LLM as a competitive quiz participant likely promotes focused, goal-driven reasoning, encouraging the retrieval of relevant knowledge and the selection of correct answers.
    
    \item reality, general: Commonsense reasoning tasks rely on everyday general knowledge. Features with semantics like ``reality'' and ``general'' may help ground the model’s responses in plausible world knowledge by activating internal representations of factual information, enabling more accurate selection of answer choices.
    
    \item think, focus: Similar to arithmetic reasoning, role-playing prompt in commonsense reasoning also involves features whose semantics are related to ``thinking'' and ``focus''. By guiding the model to pay closer attention and think more carefully, these features contribute to better reasoning performance in LLMs.

\end{itemize}

\noindent In summary, we found that in both arithmetic and commonsense reasoning tasks there exist features with semantics related to “thinking”. This suggests that one reason for the improved reasoning performance achieved through role-play prompting lies in its ability to encourage step-by-step reasoning.

\subsection{Impact of Feature Ranking in Steering}

In this section, we investigate how selecting \(k\) features from different ranking positions influences the effectiveness of steering. Specifically, we select 15 features from four different ranking ranges based on their sensitivity scores: 1–15, 6–20, 11–25, and 16–30. Features with lower sensitivity scores are nomarlly less relevant and less responsive to role-play prompts. We evaluate the performance gains of three models (Llama3.1-8B, Gemma2-2B, and Gemma2-9B) across three benchmarks (GSM8K, SVAMP, and CSQA), all under zero-shot CoT setting with fixed parameters.

\vspace{3pt}\noindent\textbf{Influence of Feature Rankings on Different Model Sizes.} As shown in Figure~\ref{fig:accuracy_gain}, model accuracy gains exhibit a decreasing trend as the selected feature rankings become lower. The most significant drop is typically observed between the top-ranked group (1–15) and the subsequent group (6–20), suggesting that the top 5 features play a particularly critical role in steering effectiveness. Specifically, Gemma2-2B exhibits sharp declines and quickly loses steering benefit beyond the first group across all benchmarks, indicating that it depends heavily on a small number of high-ranking features. In contrast, Gemma2-9B displays a more gradual decline and maintains positive or near-neutral gains even with lower-ranked features in most cases, showing stronger robustness against degraded feature quality. This trend implies that larger models with higher capacity may be more tolerant of less effective features, while smaller models are more affected by feature ranking shifts.


\vspace{3pt}\noindent\textbf{Influence of Feature Rankings on Various Datasets.} We observe varying trends in accuracy gains across the three benchmarks. In CSQA, the accuracy gains decline sharply for all models once the top 5 features are removed. This effect is particularly evident in Llama3.1-8B, whose accuracy gain drops from 7.9\% to –1.23\%, highlighting the critical role of high-quality features in commonsense reasoning tasks. In contrast, in SVAMP and GSM8K, the drop is moderate. Gemma2-9B and Llama3.1-8B maintain positive gains even when mid-ranked features are used, suggesting arithmetic reasoning tasks may tolerate larger variation in feature selection without dramatic performance loss.

\vspace{3pt}\noindent Overall, these findings confirm the effectiveness of our feature selection approach and highlight the importance of top-ranked features in steering. 

\section{Discussion}

In this section, we analyze the instability of role-play prompting and explain why SAE-based role-play steering consistently improves model performance, highlighting its advantages in controllability, stability, and generalizability across model sizes and tasks.

\vspace{3pt}\noindent\textbf{Analysis of Degraded Performance in the Role-Play Prompting Baseline.} While our steering method improves model performance in 25 out of 27 settings, role-play prompting yields accuracy gains in only 12 out of 27 settings. This indicates that simply adding a role-play prompt before the question leads to highly unstable performance, as discussed in Appendix \ref{sec:appendix_prompt_avarage}, which is consistent with findings from \citet{10.1145/3688864.3689149}. This limitation likely stems from two factors. First, our focus is on smaller models (2B to 9B), which often struggle to follow complex instructions. Second, as shown in \citet{kong2023better}, role-play prompts vary in effectiveness. Only well-crafted or specially designed prompts yield consistent gains, while simple additions are often ineffective or unstable.

It is important to note that in our role-play prompting baseline, we did not compare with the strongest prompting methods such as the specialized framework proposed in \citet{kong2023better}. Instead, we focus on comparing the performance of prompt-based methods with the steering approach utilizing features extracted from the same prompts. By extracting internal features from such prompts and applying them through our steering framework, we can further enhance the model's performance on top of what the prompts alone achieve. Compared to directly modifying the prompts, which is often unstable and highly sensitive to small changes in wording or punctuation \cite{shu-etal-2024-dont, liu2024gpt, mizrahi-etal-2024-state}, our method enables more systematic control over the model by allowing precise adjustment of steering parameters, thereby providing greater stability.

\vspace{3pt}\noindent\textbf{Underlying Reasons Behind the Consistent Effectiveness of Role-Play Steering.} While role-play prompting exhibits highly unstable performance across various evaluation settings, our SAE-based steering approach consistently delivers performance gains. The core advantage of our method lies in its controllability. Unlike prompting, where the input text is fixed once applied and cannot be adjusted dynamically, our approach allows fine-grained control over the model's behavior through tunable parameters. Moreover, prompts often include irrelevant elements such as stop words or punctuation, which may introduce noise and interfere with the model’s reasoning process, resulting in instability or performance degradation. In contrast, our SAE-based steering identifies and applies only those latent features that are strongly correlated with reasoning improvements. By precisely adjusting both the direction and strength of steering, our method amplifies beneficial components while suppressing irrelevant or harmful activations, resulting in more stable and reliable performance improvements.

Another key reason for the superiority of SAE-based steering lies in its broader applicability, particularly for smaller LLMs. These models often lack strong instruction-following capabilities and are not typically fine-tuned on role-play scenarios. As a result, appending complex prompts may exceed the model’s ability to represent information and confuse the model, ultimately harming its performance. In such cases, prompting can disrupt the model’s inherent reasoning abilities rather than enhance them. In contrast, our steering method operates directly on internal representations and does not rely on the model’s ability to interpret natural language instructions. This makes it especially well-suited for smaller models or those without extensive instruction tuning, as it can still induce targeted behavioral changes without relying on external textual guidance. Consequently, the proposed SAE-based steering proves to be a more stable intervention mechanism across different model sizes and capabilities.

\section{Conclusion}

In this work, we have introduced an efficient and interpretable steering framework that leverages features extracted from a pretrained SAE to guide LLMs toward role-specific behavior. By constructing steering vectors from highly relevant features and injecting them into the model's residual stream, our method enables fine-grained control over the model’s internal state without additional training. Extensive experiments across multiple models and reasoning benchmarks demonstrate that our approach consistently enhances model performance, yielding results on par with prompt-based role-playing. In addition, we analyze the semantic meanings of the selected features and reveal that one reason role-playing enhances the reasoning ability of LLMs is its capacity to encourage step-by-step reasoning. Lastly, we evaluate steering performance using \(k\) features selected from different ranking positions, and observe a declining trend in accuracy gains as lower-ranked features are used, validating the effectiveness of our feature selection method. Overall, our work demonstrates that steering via SAE enables implicit injection of role-playing information into the model, providing a more stable and interpretable alternative to role-play prompting for improving model reasoning ability.

\section*{Limitations}

While our method shows promising results, our experiments are primarily conducted on relatively small LLMs (2B to 9B parameters). In future, we would like to explore larger and more advanced models with stronger instruction-following capabilities, such as those exceeding 70B parameters or more recent architecture designs. Additionally, our evaluation is currently limited to three benchmarks focusing on arithmetic and commonsense reasoning. We also plan to expand the assessment to more diverse datasets spanning different domains such as logical reasoning, scientific problem-solving, and multi-modal tasks.



\bibliography{custom}

\clearpage
\appendix

\section{Implementation Details}
In this section, we provide additional details of our experiments. All experiments were run on 1 NVIDIA A100 GPU.

\subsection{Prompts}
\label{sec:appendix_prompts}
We present the role-play prompts used for steering in Table~\ref{tab:prompt}. The few-shot and one-shot CoT exemplars for arithmetic and commonsense reasoning benchmarks are shown in Tables~\ref{tab:prompt_math} and~\ref{tab:prompt_commonsense}.

\subsection{Datasets}
\label{sec:appendix_datasets}
We introduce the datasets used in our experiments below. Additional information can be found in Table~\ref{tab:dataset}.

\vspace{0.8em}
\noindent \textbf{GSM8K} (Grade School Math 8K) \cite{cobbe2021training} is a benchmark dataset for evaluating mathematical reasoning in LLMs. It contains 8,500 training examples and 1,319 test examples, each comprising a grade-school level arithmetic word problem. The problems are designed to require multi-step numerical reasoning and all answers are integers. We randomly select 1,000 samples from the training set to extract role-playing relevant features, and we evaluate the model performance on the full test set.

\vspace{0.8em}
\noindent \textbf{SVAMP} (Structured Variations of Arithmetic Math Problems) \cite{patel2021nlp} is a benchmark derived from GSM8K to evaluate the robustness of language models in arithmetic reasoning. It consists of 1,000 examples created by applying controlled variations to original problems, such as reordering and paraphrasing, challenging models to generalize beyond superficial patterns. We use the combination of full MAWPS \cite{koncel-kedziorski-etal-2016-mawps} and ASDiv-A \cite{miao2021diverse} datasets as the training set, and test model accuracy on the test set of SVAMP.

\vspace{0.8em}
\noindent \textbf{CSQA} (CommonsenseQA) \cite{talmor2018commonsenseqa} is a multiple-choice question answering dataset designed to evaluate a model’s ability to perform commonsense reasoning. It consists of 12,102 questions, each paired with five answer choices, only one of which is correct. The dataset is split into a training set (9,741 questions), a validation set (1,221 questions), and a test set (1,140 questions). Since the ground-truth answers of the test set are hidden for leaderboard evaluation, we use the validation set for model evaluation.

\subsection{Answer Extraction}
\label{sec:appendix_answer}

\begin{figure}[ht]
    \centering
    \begin{minipage}[t]{0.45\textwidth}
        \centering
        \includegraphics[width=\textwidth]{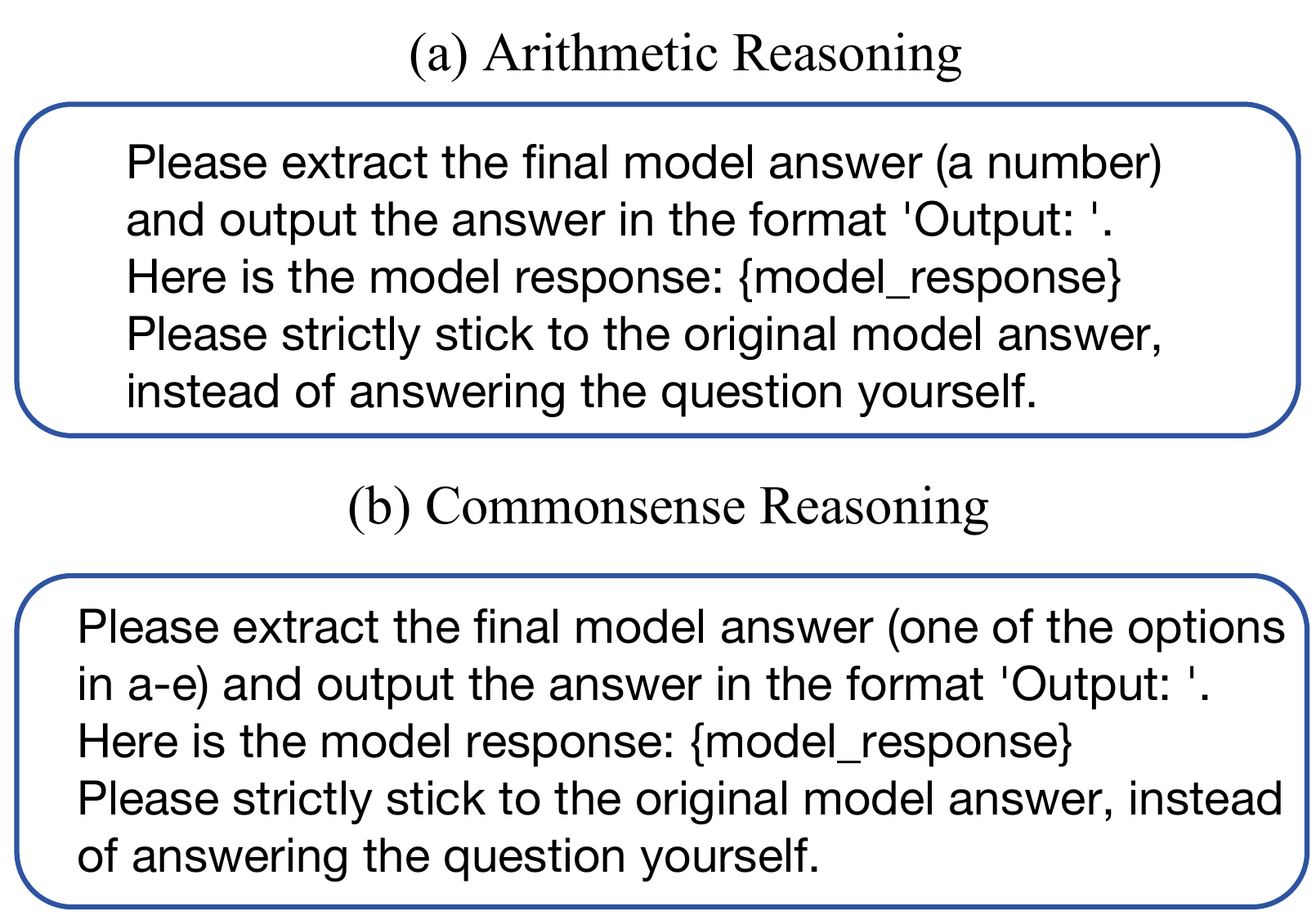}
        \captionof{figure}{Prompts used for answer extraction with GPT-4o}
        \label{fig:chatgpt}
    \end{minipage}%
    
\end{figure}

For answer generation, we set max\_new\_tokens = 150 and do\_sample = False. In the few-shot and one-shot CoT settings, since all exemplar answers appear after the keyword ``Output:'', the model learns to generate its final answer in the same way. This allows us to directly extract the answer from the text following ``Output:''. In the zero-shot CoT setting, the model does not consistently follow a fixed output format, making answer extraction less reliable. To address this, we use GPT-4o \cite{hurst2024gpt} to extract answers from the model’s free-form outputs. The prompts used for this extraction process are shown in Figure~\ref{fig:chatgpt}. We apply greedy decoding with temperature = 0 to obtain deterministic results. Since GPT-4o is a powerful model, it can occasionally answer the question itself when the original model fails to produce a clear response. To maintain a fair evaluation of model performance, we hide the original question and provide only the model’s response as input.

\begin{table*}[ht]
\centering
   \begin{tabular}{p{3.5cm}p{3cm}p{1.5cm}p{1.5cm}p{2cm}}
\toprule
\toprule
Dataset & Answer Format                     & \multicolumn{1}{l}{$N_{\text{train}}$} & \multicolumn{1}{l}{$N_{\text{test}}$} & \multicolumn{1}{c}{License}\\ \midrule
GSM8K   & \multicolumn{1}{l}{arabic number} & 7,473                                   & 1,319     &MIT License                          \\ 
SVAMP   & arabic number                     & 3,138                                   & 1,000            & MIT License                      \\ 
CommonsenseQA    & \multicolumn{1}{l}{option(A-E)}   & 9,741                                   & 1,221                   & Unspecified               \\ \bottomrule
\bottomrule
\end{tabular}%

\caption{Information about the evaluated benchmarks. $N_{\text{train}}$ denotes the size of the training set, while $N_{\text{test}}$ denotes the number of test samples.}
\label{tab:dataset}
\end{table*}

\section{Performance of Role-Play Prompting}

\label{sec:appendix_prompt_avarage}

Table~\ref{tab:avarage_prompt} presents model accuracy across five role-playing prompts, along with their average and standard deviation. The accuracy variation across the five prompts increases as the number of few-shot exemplars decreases, particularly in the zero-shot CoT setting, indicating the instability of the prompt-based role-playing.

\section{Hyperparameters of SAE Steering}

\label{sec:appendix_hyperparameters}

For each dataset, we perform a grid search over the first 100 test questions to identify relatively optimal parameter combinations. We report all the hyperparameters used for steering across different evaluation settings on the three benchmarks. As the number of features \(k\) selected for steering is fixed at 15 in all experiments, we focus on comparing the remaining hyperparameters (\(\theta\), \(\beta\), and \(\lambda\)) in Table~\ref{tab:hyperparameter}. Specifically, \(\theta\) is the activation threshold used to determine whether a feature is considered active; \(\beta\) is a weighting coefficient that balances the contribution of activation frequency and strength in feature selection; and \(\lambda\) is a scaling factor that controls the strength of the steering vector when injected into the model's residual stream.

\begin{table}[ht]
\centering

\begin{subtable}{\columnwidth}
\centering
\resizebox{\columnwidth}{!}{%
\begin{tabular}{l|ccc|ccc|ccc}
\toprule
\toprule
            & \multicolumn{3}{c|}{Few-shot-CoT} & \multicolumn{3}{c|}{One-shot-CoT} & \multicolumn{3}{c}{Zero-shot-CoT} \\
\midrule
            & \(\theta\) & \(\beta\) & \(\lambda\) & \(\theta\) & \(\beta\) & \(\lambda\) & \(\theta\) & \(\beta\) & \(\lambda\) \\
\midrule
Llama3.1-8B & 0.2 & 3  & 5  & 0.2 & 3 & 3  & 0.2 & 3  & 3 \\
Gemma2-2B   & 0.2 & 3  & 8  & 0.2 & 5 & 14 & 0.2 & 5  & 13 \\
Gemma2-9B   & 0.2 & 5  & 11 & 0.2 & 10 & 5 & 0.2 & 10 & 6 \\
\bottomrule
\bottomrule
\end{tabular}%
}
\caption{GSM8K}
\end{subtable}

\vspace{1em}

\begin{subtable}{\columnwidth}
\centering
\resizebox{\columnwidth}{!}{%
\begin{tabular}{l|ccc|ccc|ccc}
\toprule
\toprule
            & \multicolumn{3}{c|}{Few-shot-CoT} & \multicolumn{3}{c|}{One-shot-CoT} & \multicolumn{3}{c}{Zero-shot-CoT} \\
\midrule
            & \(\theta\) & \(\beta\) & \(\lambda\) & \(\theta\) & \(\beta\) & \(\lambda\) & \(\theta\) & \(\beta\) & \(\lambda\) \\
\midrule
Llama3.1-8B & 0.2 & 3  & 4  & 0   & 5 & 4  & 0.3 & 3  & 4 \\
Gemma2-2B   & 0.2 & 3  & 5  & 0.2 & 3 & 10 & 0.2 & 3  & 10 \\
Gemma2-9B   & 0.2 & 10 & 20 & 0.2 & 10 & 30 & 0.2 & 10 & 30 \\
\bottomrule
\bottomrule
\end{tabular}%
}
\caption{SVAMP}
\end{subtable}

\vspace{1em}

\begin{subtable}{\columnwidth}
\centering
\resizebox{\columnwidth}{!}{%
\begin{tabular}{l|ccc|ccc|ccc}
\toprule
\toprule
            & \multicolumn{3}{c|}{Few-shot-CoT} & \multicolumn{3}{c|}{One-shot-CoT} & \multicolumn{3}{c}{Zero-shot-CoT} \\
\midrule
            & \(\theta\) & \(\beta\) & \(\lambda\) & \(\theta\) & \(\beta\) & \(\lambda\) & \(\theta\) & \(\beta\) & \(\lambda\) \\
\midrule
Llama3.1-8B & 0.2 & 3  & 10 & 0.3   & 3 & 10  & 0.3 & 3  & 10 \\
Gemma2-2B   & 0.3 & 4  & 5  & 0.3 & 15 & 5 & 0.3 & 15  & 5 \\
Gemma2-9B   & 0.2 & 3 & 35 & 0.2 & 3 & 35 & 0.2 & 3 & 10 \\
\bottomrule
\bottomrule
\end{tabular}%
}
\caption{CSQA}
\end{subtable}

\vspace{0.8em}
\caption{Hyperparameters used by different models across all evaluation settings on the GSM8K, SVAMP, and CSQA benchmarks.}
\label{tab:hyperparameter}
\end{table}

Among the three models, Llama3.1-8B is the most sensitive to activation perturbation, and small changes in parameter values can lead to significant differences in steering performance. In contrast, the Gemma-2 family models, particularly Gemma2-9B, is less sensitive to activation modification. Among the three hyperparameters, \(\beta\) and \(\lambda\) have the greatest impact on steering performance. For models that are more robust to activation intervention, such as Gemma2-9B, larger values of \(\beta\) and \(\lambda\) (around 10) are typically preferred, and parameter tuning can be performed with larger step sizes (e.g., increments of 5). In contrast, for more steering-sensitive models like Llama3.1-8B, smaller values of \(\beta\) and \(\lambda\) (around 5) tend to yield better results, and finer-grained tuning with smaller steps (e.g., increments of 1) is recommended. In general, \(\theta\) has minimal influence on steering performance and can be set to a default value (e.g., 0.2 or 0.3) to maintain the effectiveness of the steering vectors.

\begin{table*}[ht]
 \centering
   \begin{tabular}{p{5cm}p{10cm}}
                                    \vspace{4pt}\\
   \toprule
                                       & \multicolumn{1}{c}{\textbf{Role-Play Prompt}} 

                                       \\
\toprule
\multirow{15}{*}{\textbf{Arithmetic Reasoning}}  & As a highly qualified mathematics teacher, you excel at solving problems systematically and explaining solutions with clarity. I am your student, eager to learn. Please solve the following problem:                                                                      \\
& As an excellent mathematics teacher, you always guide your students correctly through math problems. I am one of your students, eager to learn. Please answer the following question:                                                                                      \\
& As a respected mathematics professor with deep expertise in solving complex problems, you are known for your clarity and precision. I am your student and need help. Please solve the following question for me:                                                           \\
& As a world-renowned mathematics teacher, you are highly skilled at solving problems precisely and explaining them effectively. I am your student, struggling with a question. Please solve the following task for me:                                                      \\
& As a mathematics expert with strong problem-solving skills, you are deeply trusted by your students. I am one of them and need your help. Please solve the following problem for me:                                                                                       \vspace{2pt}\\
\midrule
\multirow{20}{*}{\textbf{Commonsense Reasoning}} & You are now a contestant in a general knowledge quiz and are always able to answer all kinds of common sense questions accurately. I am the host of the contest, and the final round is about to begin. Let’s kick things off with your first question:                    \\
& Please take on the role of a contestant in a general knowledge competition, capable of answering all types of common sense questions correctly. The contest has reached the final stage, and I am the moderator. Here comes your first challenge:                          \\
& From this point on, you will appear as a participant in a general knowledge quiz, and you must respond accurately to every common sense question. I am the host of this final round, and the contest is about to start. Let’s begin with the first question:               \\
& Imagine that you are now a contestant in a general knowledge competition, able to correctly answer any question involving common sense. The final is about to begin, and I will be hosting the match. Now, let’s see how you do with the first question:                   \\
& You will take on the role of a contestant in a general knowledge quiz, equipped with the ability to answer all types of common sense questions precisely. As the host, I announce that the final round is about to commence. Let’s start the game with the first question:\vspace{2pt}\\
\bottomrule

\end{tabular}%

\caption{Role-play prompts used in arithmetic and commonsense reasoning benchmarks}
\label{tab:prompt}
\end{table*}

\begin{table*}[t]
\centering
\begin{tabular}{p{0.95\textwidth}} 
\toprule
\textbf{One-shot exemplar} \\
\toprule
\textbf{Q}: Michael had 58 golf balls. On Tuesday, he lost 23 golf balls. On Wednesday, he lost 2 more. How many golf balls did he have at the end of Wednesday? \\
\textbf{A}: Let's think step by step. Michael started with 58 golf balls. After losing 23 on Tuesday, he had 58 - 23 = 35. After losing 2 more, he had 35 - 2 = 33. \\
\textbf{Output}: 33 \\
\\[-0.7em]
\toprule
\textbf{Few-shot exemplars} \\
\toprule
\textbf{Q}: Jason had 20 lollipops. He gave Denny some lollipops. Now Jason has 12 lollipops. How many lollipops did Jason give to Denny? \\
\textbf{A}: Let's think step by step. Jason started with 20 lollipops. Then he had 12 after giving some to Denny. So he gave Denny 20 - 12 = 8. \\
\textbf{Output}: 8 \\
\\[-0.5em]
\textbf{Q}: Leah had 32 chocolates and her sister had 42. If they ate 35, how many pieces do they have left in total? \\
\textbf{A}: Let's think step by step. Originally, Leah had 32 chocolates. Her sister had 42. So in total they had 32 + 42 = 74. After eating 35, they had 74 - 35 = 39. \\
\textbf{Output}: 39 \\
\\[-0.5em]
\textbf{Q}: There were nine computers in the server room. Five more computers were installed each day, from Monday to Thursday. How many computers are now in the server room? \\
\textbf{A}: Let's think step by step. There were originally 9 computers. For each of 4 days, 5 more computers were added. So 5 * 4 = 20 computers were added. 9 + 20 = 29. \\
\textbf{Output}: 29 \\
\\[-0.5em]
\textbf{Q}: Olivia has \$23. She bought five bagels for \$3 each. How much money does she have left? \\
\textbf{A}: Let's think step by step. Olivia had 23 dollars. 5 bagels for 3 dollars each will be 5 x 3 = 15 dollars. So she has 23 - 15 dollars left. 23 - 15 = 8. \\
\textbf{Output}: 8 \\
\bottomrule
\end{tabular}
\caption{One-shot and few-shot exemplars with chain-of-thought for arithmetic reasoning benchmarks.}
\label{tab:prompt_math}
\end{table*}

\begin{table*}[t]
\centering
\begin{tabular}{p{0.95\textwidth}} 
\toprule
\textbf{One-shot exemplar} \\
\toprule
\textbf{Q}: What home entertainment equipment requires cable? Answer Choices: (a) radio shack (b) substation (c) television (d) cabinet \\
\textbf{A}: Let's think step by step. The answer must require cable. Of the above choices, only television requires cable. \\
\textbf{Output}: (c) \\
\\[-0.7em]
\toprule
\textbf{Few-shot exemplars} \\
\toprule
\textbf{Q}: Where do you put your grapes just before checking out? Answer Choices: (a) mouth (b) grocery cart (c)super market (d) fruit basket (e) fruit market \\
\textbf{A}: Let's think step by step. The answer should be the place where grocery items are placed before checking out. Of the above choices, grocery cart makes the most sense for holding grocery items. \\
\textbf{Output}: (b) \\
\\[-0.5em]
\textbf{Q}: Google Maps and other highway and street GPS services have replaced what? Answer Choices: (a) united states (b) mexico (c) countryside (d) atlas \\
\textbf{A}: Let's think step by step. The answer must be something that used to do what Google Maps and GPS services do, which is to give directions. Of the above choices, only atlases are used to give directions. \\
\textbf{Output}: (d) \\
\\[-0.5em]
\textbf{Q}: Before getting a divorce, what did the wife feel who was doing all the work? Answer Choices: (a) harder (b) anguish (c) bitterness (d) tears (e) sadness \\
\textbf{A}: Let's think step by step. The answer should be the feeling of someone getting divorced who was doing all the work. Of the above choices, the closest feeling is bitterness. \\
\textbf{Output}: (c)  \\
\\[-0.5em]
\textbf{Q}: What home entertainment equipment requires cable? Answer Choices: (a) radio shack (b) substation (c) television (d) cabinet \\
\textbf{A}: Let's think step by step. The answer must require cable. Of the above choices, only television requires cable. \\
\textbf{Output}: (c) \\
\bottomrule
\end{tabular}
\caption{One-shot and few-shot exemplars with chain-of-thought for commonsense reasoning benchmarks.}
\label{tab:prompt_commonsense}
\end{table*}

\begin{table*}
\centering
\resizebox{\textwidth}{!}{%
\begin{tabular}{c|c|c|c|c|c|c|c|c|c}
\toprule[1pt]
\toprule[1pt]
Model                                            & Task                   & \multicolumn{1}{l|}{Evaluation Setting} & Prompt 1 & Prompt 2 & Prompt 3 & Prompt 4 & Prompt 5 & \multicolumn{1}{l|}{Average} & Standard Deviation \\
\midrule
\multirow{9}{*}{Gemma2-2B}                       & \multirow{3}{*}{GSM8K} & 0-shot                                  & 24.26    & 17.66    & 19.18    & 23.28    & 23.20    & 21.52                        & \textbf{2.60}               \\
                                                 &                        & 1-shot                                  & 18.20    & 15.69    & 16.45    & 16.45    & 16.60    & 16.68                        & 0.82               \\
                                                 &                        & 4-shot                                  & 27.45    & 26.91    & 27.90    & 26.91    & 27.82    & 27.40                        & 0.43               \\
                                                 \cmidrule{2-10}
                                                 & \multirow{3}{*}{SVAMP} & 0-shot                                  & 53.30    & 52.30    & 44.10    & 55.10    & 48.20    & 50.60                        & \textbf{3.96}               \\
                                                 &                        & 1-shot                                  & 38.90    & 39.10    & 39.10    & 38.70    & 39.20    & 39.00                        & 0.18               \\
                                                 &                        & 4-shot                                  & 44.70    & 45.40    & 44.20    & 45.50    & 43.70    & 44.70                        & 0.69               \\
                                                 \cmidrule{2-10}
                                                 & \multirow{3}{*}{CSQA}  & 0-shot                                  & 20.72    & 22.11    & 20.72    & 22.11    & 25.88    & 22.31                        & \textbf{1.89}               \\
                                                 &                        & 1-shot                                  & 52.91    & 49.55    & 54.22    & 51.19    & 50.20    & 51.61                        & 1.73               \\
                                                 &                        & 4-shot                                  & 58.97    & 59.54    & 60.11    & 58.89    & 58.97    & 59.30                        & 0.47               \\
                                                 \midrule
\multicolumn{1}{l|}{\multirow{9}{*}{Gemma2-9B}} & \multirow{3}{*}{GSM8K} & 0-shot                                  & 44.73    & 44.12    & 44.28    & 49.73    & 43.97    & 45.37                        & \textbf{2.20}               \\
\multicolumn{1}{l|}{}                           &                        & 1-shot                                  & 55.88    & 59.51    & 55.34    & 56.18    & 56.03    & 56.59                        & 1.49               \\
\multicolumn{1}{l|}{}                           &                        & 4-shot                                  & 65.20    & 65.05    & 64.82    & 65.05    & 65.50    & 65.13                        & 0.22               \\
\cmidrule{2-10}
\multicolumn{1}{l|}{}                           & \multirow{3}{*}{SVAMP} & 0-shot                                  & 67.80    & 65.80    & 63.50    & 75.30    & 73.90    & 69.26                        & \textbf{4.59}               \\
\multicolumn{1}{l|}{}                           &                        & 1-shot                                  & 70.60    & 75.70    & 71.70    & 70.90    & 74.60    & 72.70                        & 2.06               \\
\multicolumn{1}{l|}{}                           &                        & 4-shot                                  & 78.70    & 77.30    & 79.40    & 79.30    & 77.80    & 78.50                        & 0.83               \\
\cmidrule{2-10}
\multicolumn{1}{l|}{}                           & \multirow{3}{*}{CSQA}  & 0-shot                                  & 39.56    & 40.62    & 43.33    & 46.44    & 44.39    & 42.87                        & \textbf{2.50}               \\
\multicolumn{1}{l|}{}                           &                        & 1-shot                                  & 72.07    & 71.99    & 73.46    & 71.25    & 73.30    & 72.42                        & 0.84               \\
\multicolumn{1}{l|}{}                           &                        & 4-shot                                  & 78.05    & 77.31    & 78.13    & 77.81    & 78.05    & 77.87                        & 0.30               \\
\midrule
\multirow{9}{*}{Llama3.1-8B}                     & \multirow{3}{*}{GSM8K} & 0-shot                                  & 35.33    & 38.13    & 36.92    & 33.97    & 34.95    & 35.86                        & \textbf{1.48}               \\
                                                 &                        & 1-shot                                  & 40.18    & 41.85    & 38.59    & 40.03    & 41.47    & 40.42                        & 1.16               \\
                                                 &                        & 4-shot                                  & 52.39    & 53.75    & 53.15    & 53.37    & 54.51    & 53.43                        & 0.70               \\
                                                 \cmidrule{2-10}
                                                 & \multirow{3}{*}{SVAMP} & 0-shot                                  & 46.70    & 54.60    & 47.30    & 52.60    & 52.90    & 50.82                        & \textbf{3.20}               \\
                                                 &                        & 1-shot                                  & 53.50    & 54.90    & 55.40    & 55.90    & 54.90    & 54.92                        & 0.80               \\
                                                 &                        & 4-shot                                  & 64.60    & 65.40    & 63.90    & 64.30    & 64.50    & 64.54                        & 0.49               \\
                                                 \cmidrule{2-10}
                                                 & \multirow{3}{*}{CSQA}  & 0-shot                                  & 20.07    & 16.22    & 18.59    & 19.33    & 27.27    & 20.29                        & \textbf{3.72}               \\
                                                 &                        & 1-shot                                  & 65.19    & 64.95    & 65.19    & 65.03    & 64.05    & 64.88                        & 0.43               \\
                                                 &                        & 4-shot                                  & 73.14    & 71.83    & 73.71    & 72.24    & 73.22    & 72.83                        & 0.69    \\
                                            \bottomrule[1pt]
                                            \bottomrule[1pt]
\end{tabular}%
}
\caption{Accuracy of each model evaluated with five distinct role-playing prompts on three datasets under different evaluation settings. The final two columns report the average accuracy and standard deviation across the five role-play prompts. The largest standard deviation in three evaluation settings per group is marked in \textbf{bold}.}
\label{tab:avarage_prompt}
\end{table*}

\section{Examples of Steering Effect}
\label{sec: steering_example}

\begin{figure*}[ht]
    \centering
    \includegraphics[width=0.9\textwidth]{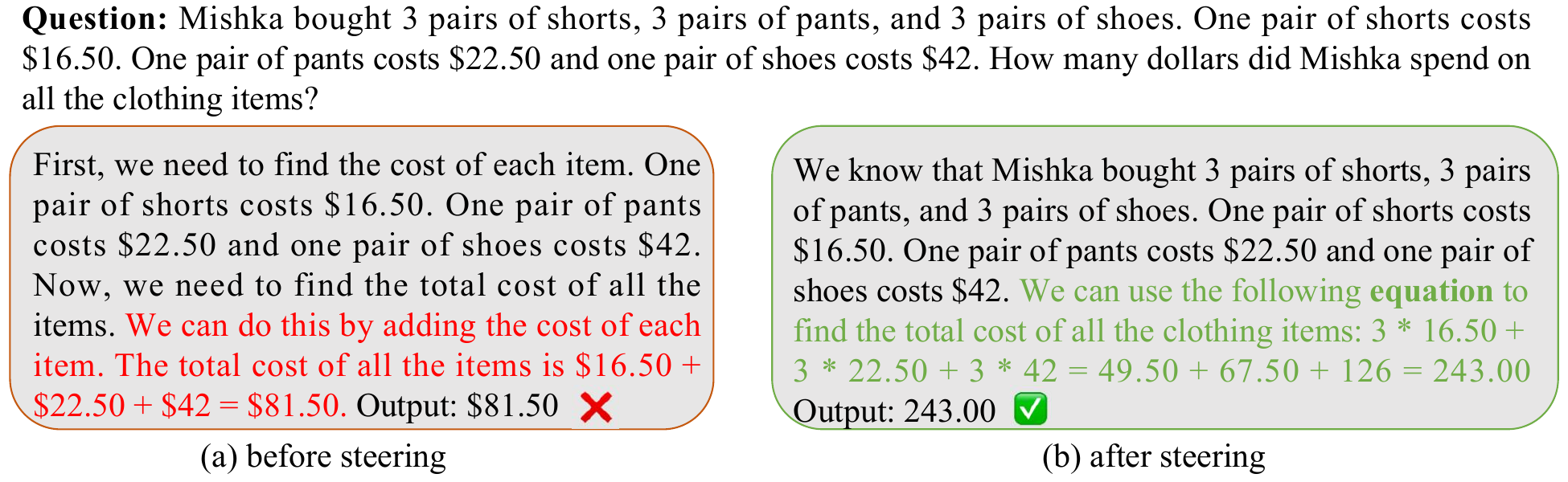}
    \includegraphics[width=0.9\textwidth]{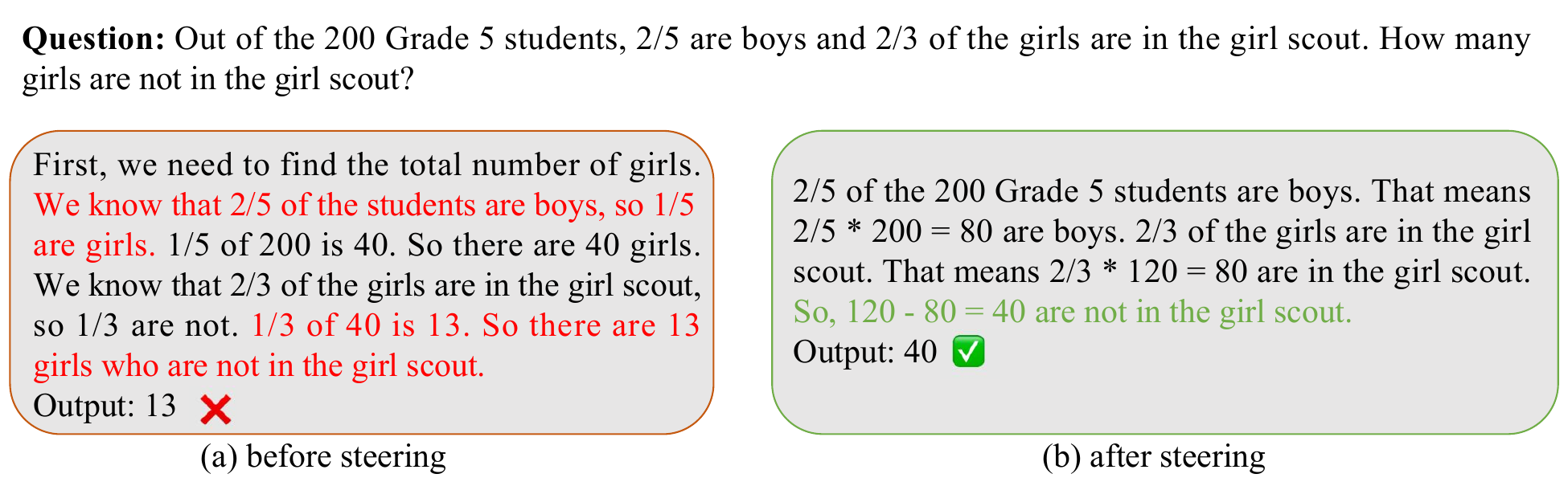}
    \includegraphics[width=0.9\textwidth]{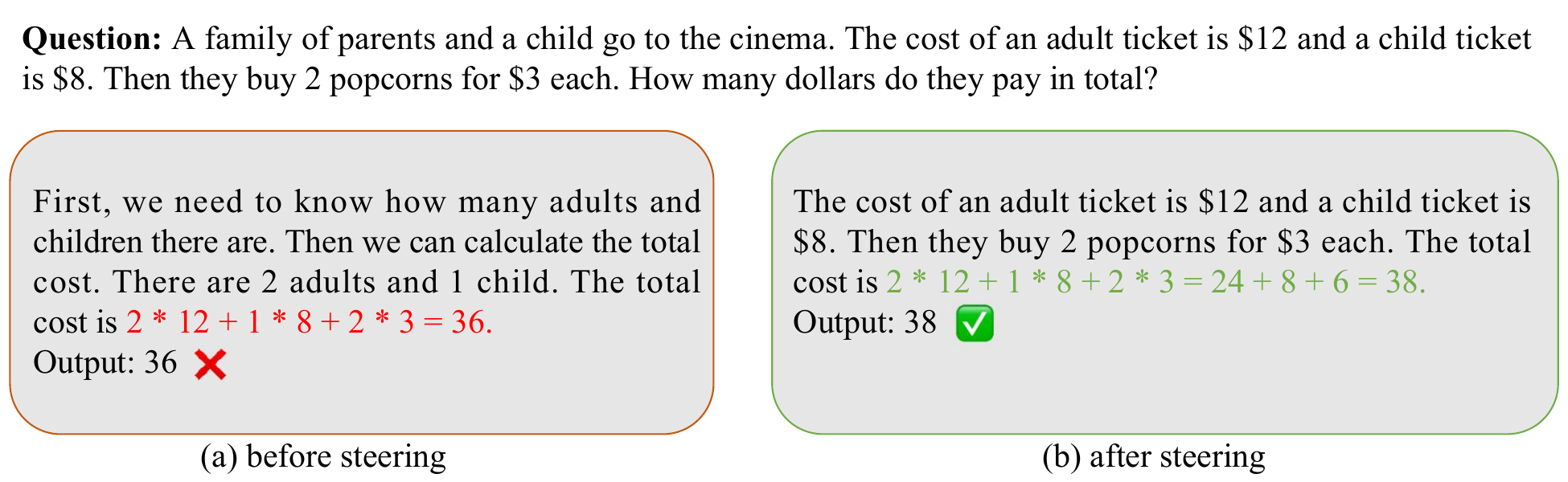}
    \includegraphics[width=0.9\textwidth]{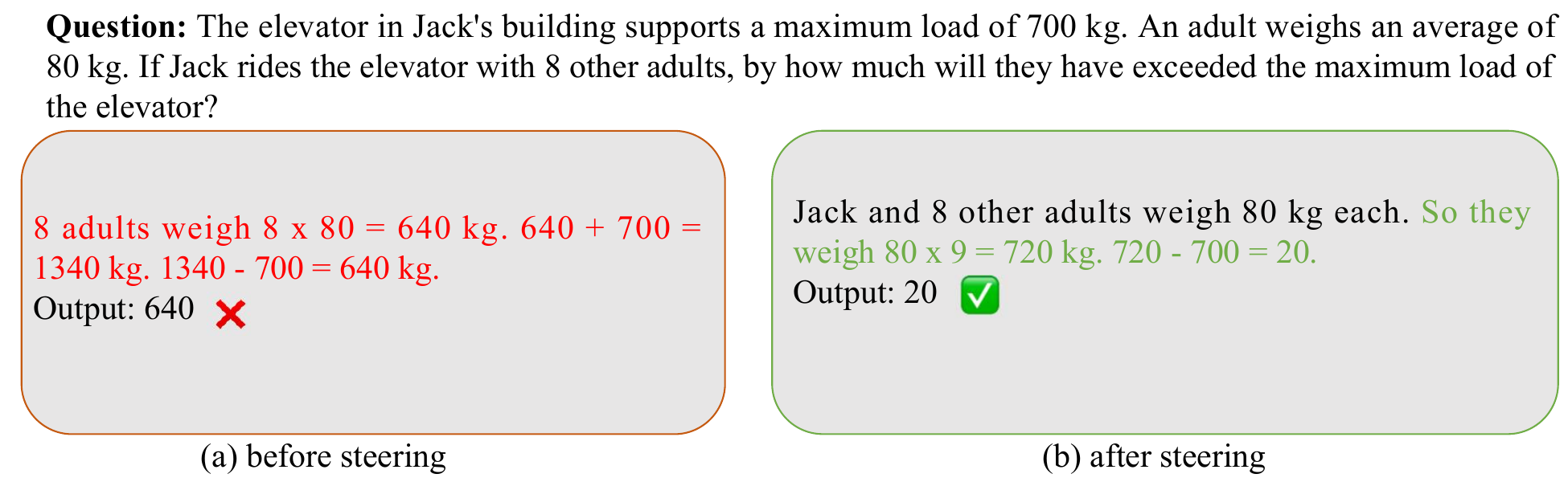}
    \includegraphics[width=0.9\textwidth]{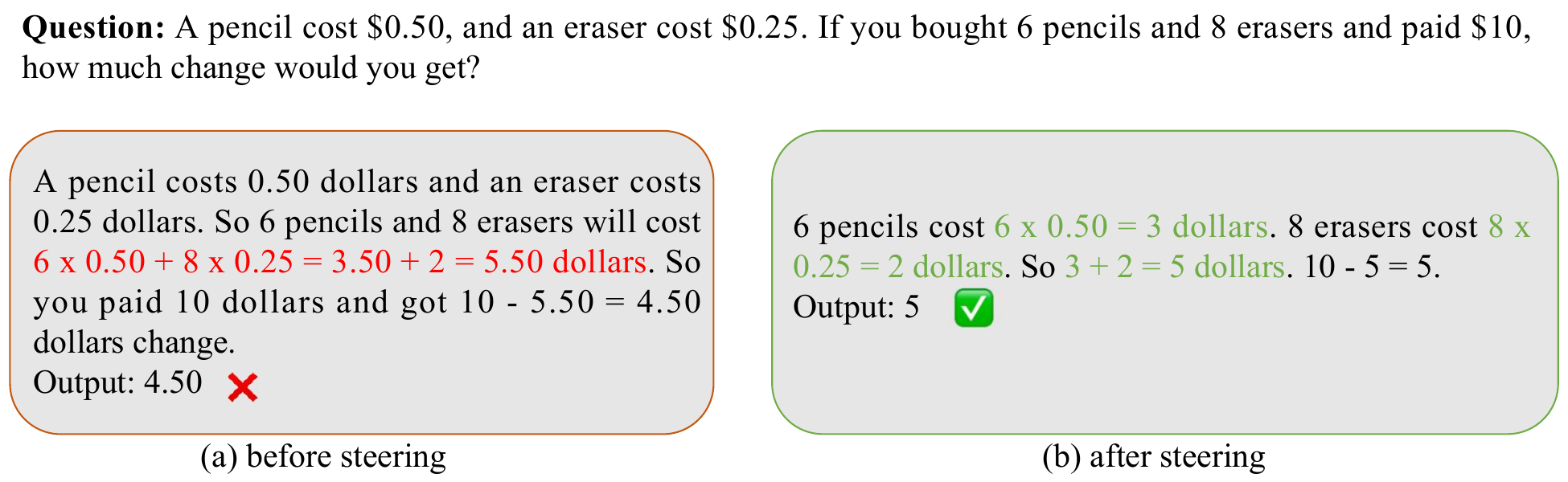}
    \caption{Steering examples from the GSM8K dataset.}
    \label{fig:gsm8k_examples}
\end{figure*}

\begin{figure*}[ht]
    \centering
    \includegraphics[width=0.9\textwidth]{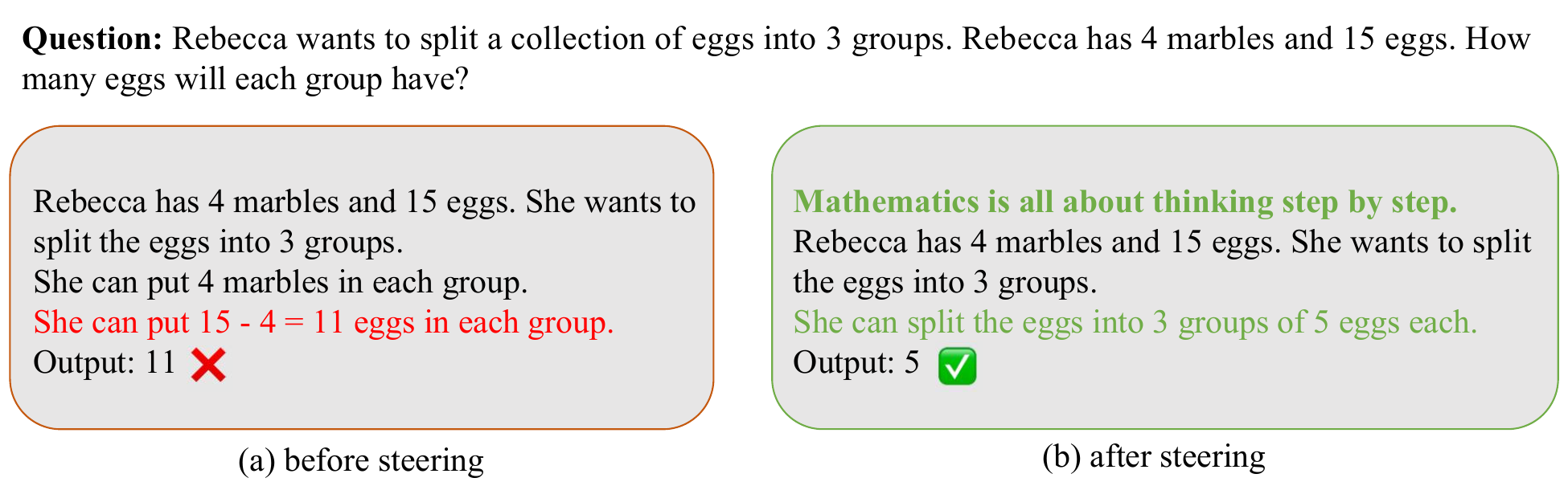}
    \includegraphics[width=0.9\textwidth]{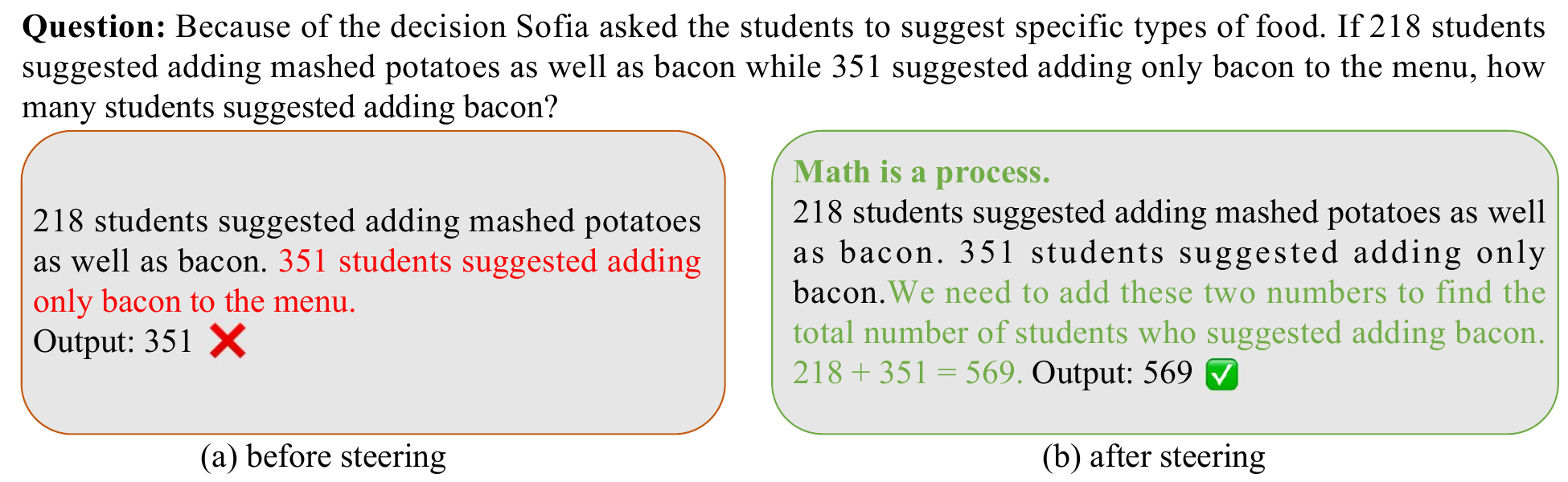}
    \includegraphics[width=0.9\textwidth]{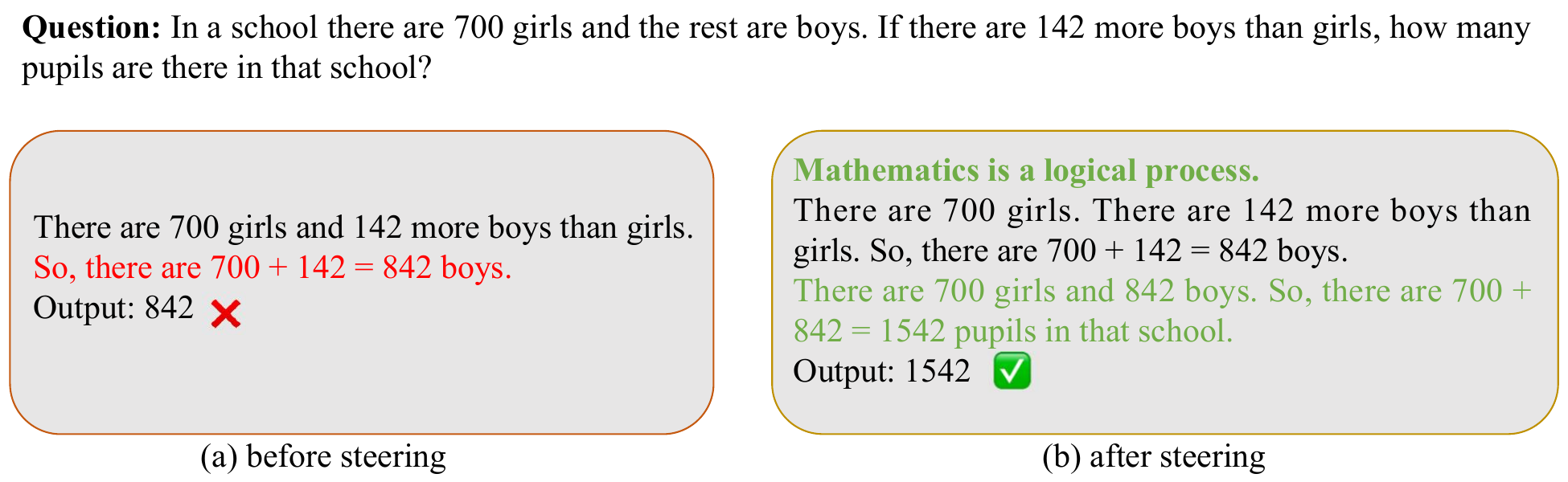}
    \includegraphics[width=0.9\textwidth]{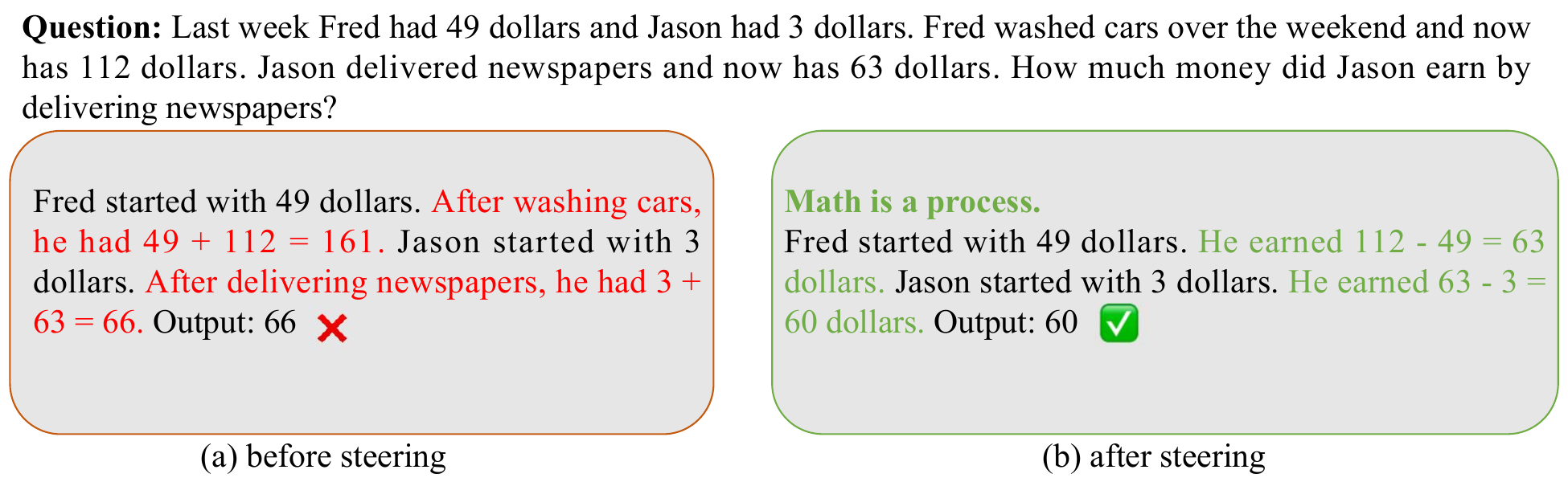}
    \includegraphics[width=0.9\textwidth]{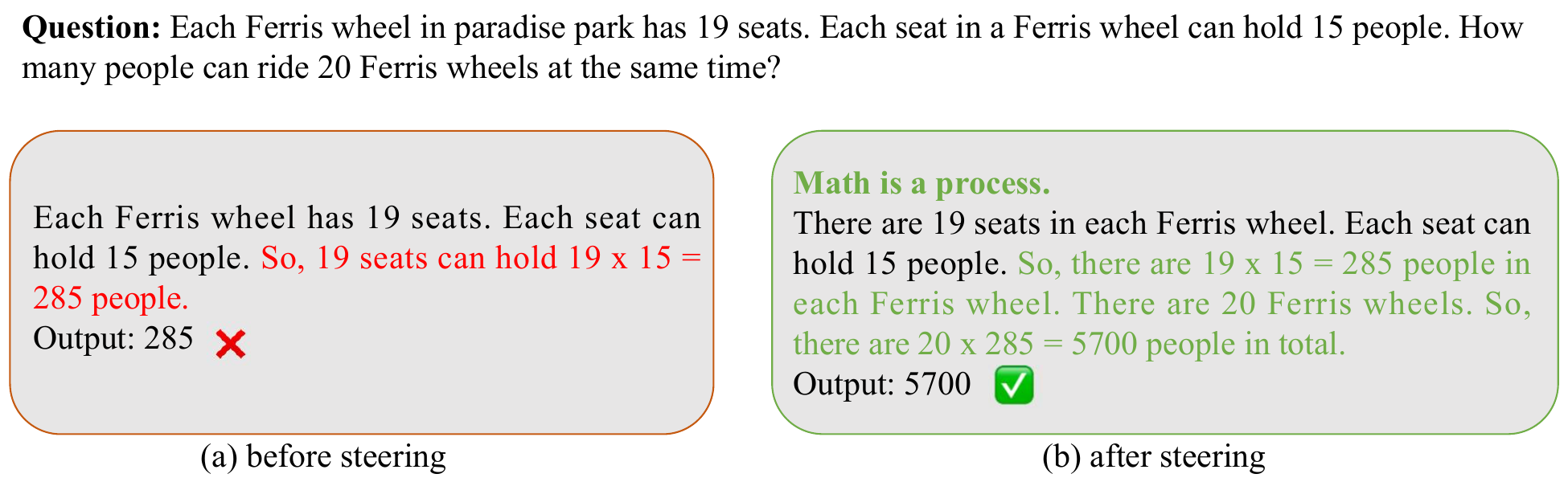}
    
    \caption{Steering examples from the SVAMP dataset.}
    \label{fig:svamp_examples}
\end{figure*}

\begin{figure*}[ht]
    \centering
    \includegraphics[width=0.9\textwidth]{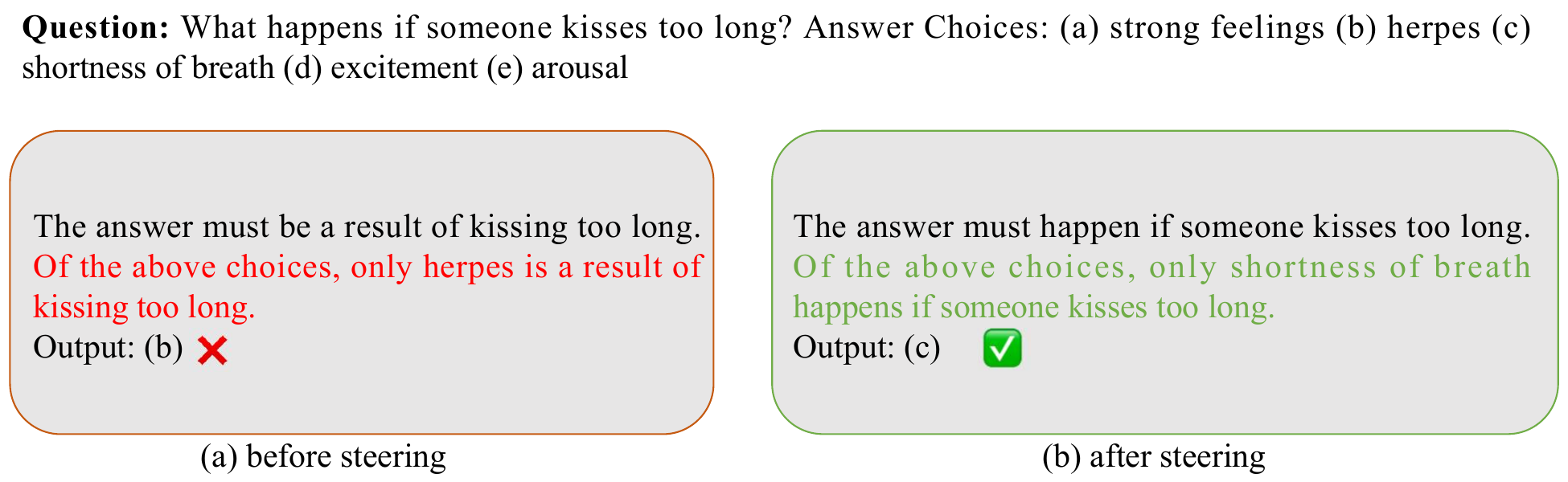}
    \includegraphics[width=0.9\textwidth]{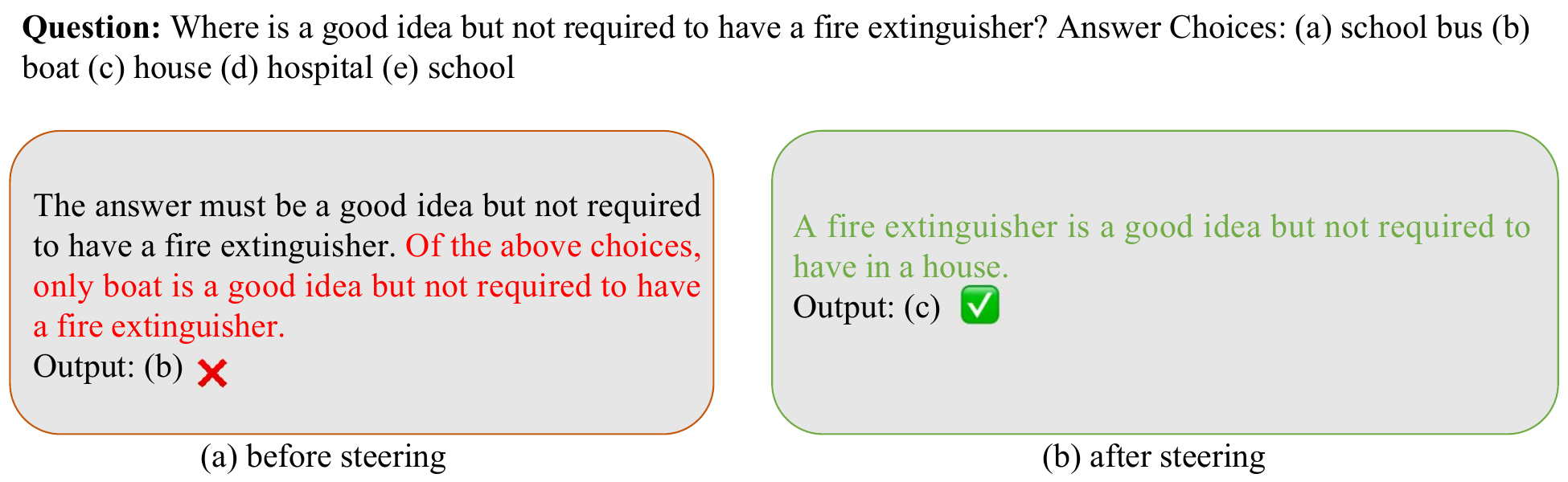}
    \includegraphics[width=0.9\textwidth]{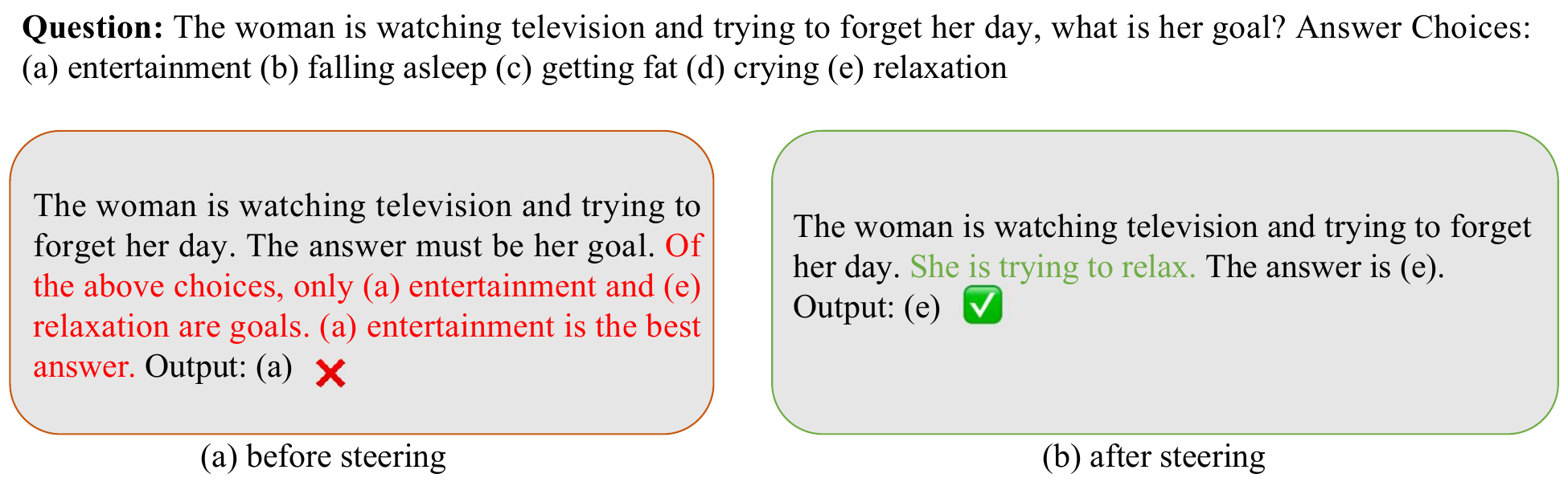}
    \includegraphics[width=0.9\textwidth]{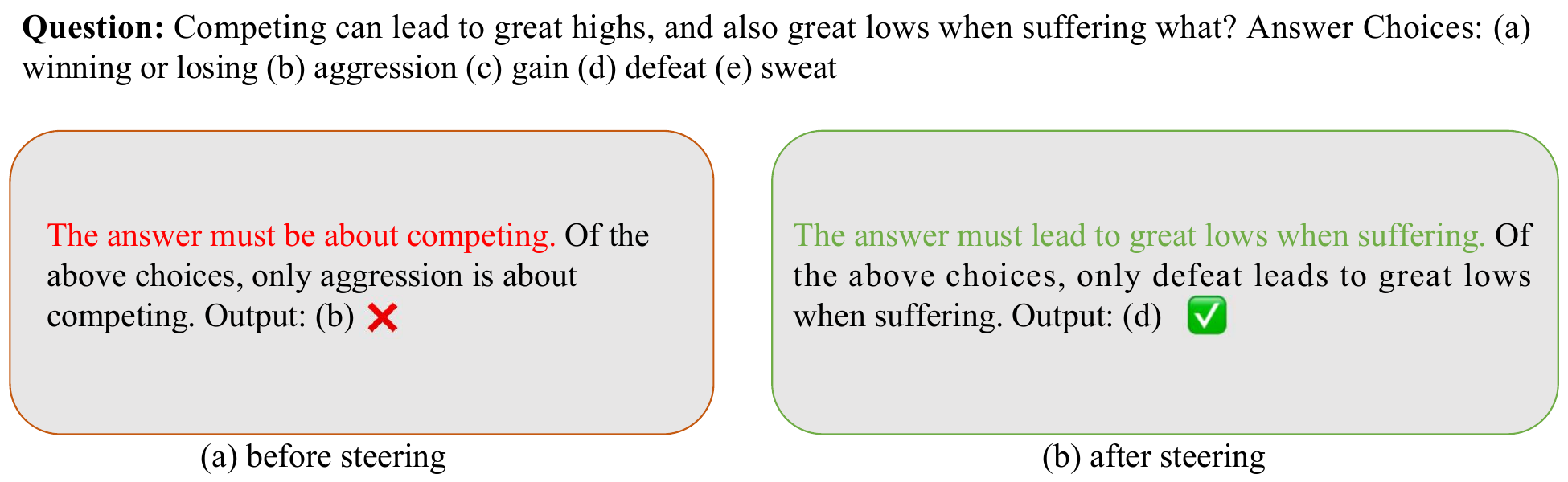}
    \includegraphics[width=0.9\textwidth]{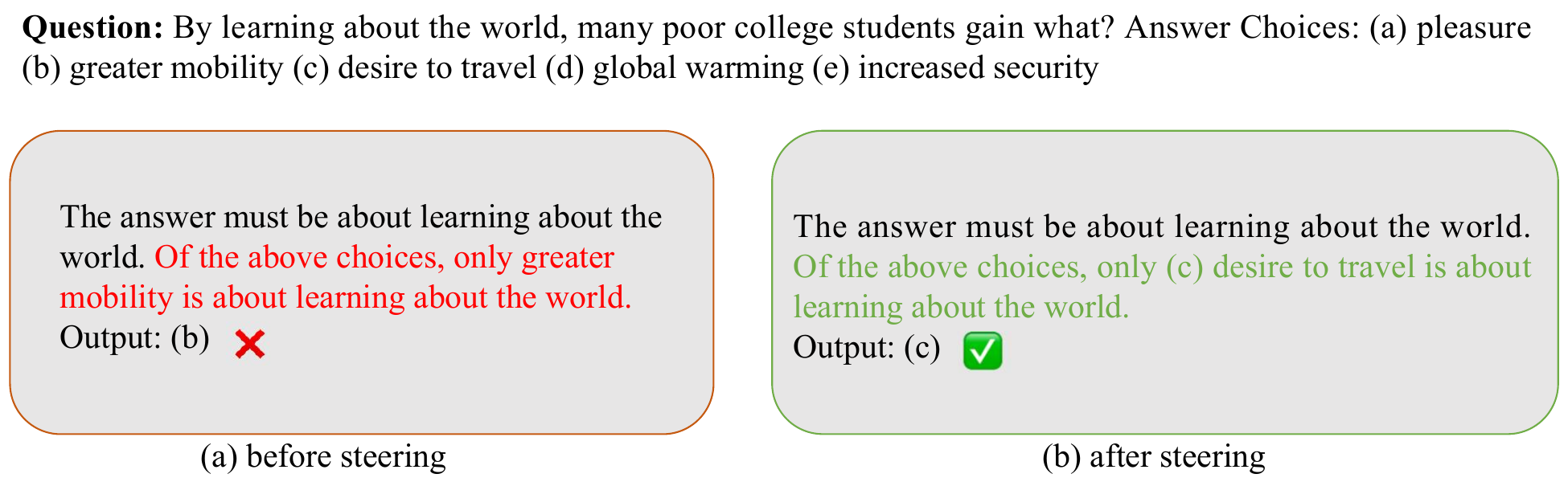}
    
    \caption{Steering examples from the CSQA dataset.}
    \label{fig:csqa_examples}
\end{figure*}

We select five representative examples from each dataset to illustrate the effectiveness of our role-playing steering method (see Figure~\ref{fig:gsm8k_examples}, \ref{fig:svamp_examples}, and \ref{fig:csqa_examples}).

In GSM8K, the steered model shows improved arithmetic reasoning ability by breaking down the calculation process into multiple explicit steps. Prior to steering, the model often omits key details and tends to perform computations using only partial information from the first half of the question, resulting in incorrect answers. After steering, the model demonstrates a deeper understanding of the problem structure and can effectively incorporate all relevant information presented in the question, thereby producing more accurate solutions.

In SVAMP, steering enhances the model’s ability to avoid being distracted by misleading information added to the problem. SVAMP questions are designed to introduce subtle variations and irrelevant cues to challenge model robustness.
Without steering, the model is more likely to be affected by these interventions and frequently produces incorrect answers. In contrast, the steered model generates reasoning paths containing tokens such as ``mathematics'', ``logical'', and ``think step by step'', indicating stronger alignment with systematic problem-solving behavior. The model is better able to filter out misleading information and focus on task-relevant content when deriving answers.

In CSQA, the steered model can more effectively leverage general world knowledge for commonsense reasoning. Compared to the original model, which may depend on surface-level clues or misinterpret the question, the steered model produces answers that reflect a more coherent understanding of real-world concepts and causal relationships. This suggests that our method not only enhances reasoning in numerical tasks but also facilitates more faithful retrieval and application of factual knowledge in commonsense question answering.

\end{document}